\documentclass[10pt,twocolumn,letterpaper]{article}
\usepackage{wacv}
\usepackage[mode=buildnew]{standalone}
\usepackage{times}
\usepackage{epsfig}
\usepackage{graphicx}
\usepackage{amsmath}
\usepackage{amssymb}
\usepackage{booktabs}
\usepackage{comment}
\usepackage{todonotes}
\usepackage[utf8]{inputenc}
\usepackage{float}
\usepackage{amsfonts}
\usepackage{siunitx}
\usepackage{adjustbox}
\usepackage{lipsum}
\usepackage{multirow}
\usepackage{tabularx}
\usepackage{subcaption}

\usepackage{pifont}% http://ctan.org/pkg/pifont
\usepackage{caption}
\usepackage[pagebackref,breaklinks,colorlinks]{hyperref}
\usepackage[capitalize]{cleveref}

\usetikzlibrary{shapes.geometric}
\newcommand*{\subb}[1]{\ensuremath{_{\mathrm{#1}}}}
\newcommand*{\supp}[1]{\ensuremath{^{\mathrm{#1}}}}

\crefname{section}{Sec.}{Secs.}
\Crefname{section}{Section}{Sections}
\Crefname{table}{Table}{Tables}
\crefname{table}{Tab.}{Tabs.}

\usetikzlibrary{pgfplots.groupplots}

\begin{document}

%%%%%%%%% TITLE
\title{MatChA: Cross-Algorithm Matching with Feature Augmentation}

\author{
 Paula Carbó Cubero$^{1,2}$~~~
Alberto Jaenal Gálvez$^{2}$~~~
André Mateus$^{2}$~~~
José Araújo$^{2}$~~~
Patric Jensfelt$^{1}$~~~
\\
\small{
$^{1}$KTH Royal Institute of Technology \ 
$^{2}$Ericsson Research}
\\
{\tt\small paulacc@kth.se}
}

% Enter the institutions
% \addinstitution{Name\\Address}
%\addinstitution{
% Ericsson Research\\
% Stockholm, Sweden
%}
%\addinstitution{
% KTH Royal Institute of Technology\\
% Stockholm, Sweden
%}

\maketitle
%\thispagestyle{empty}

%%%%%%%%% ABSTRACT
\begin{abstract}
   State-of-the-art methods fail to solve visual localization in scenarios where different devices use different sparse feature extraction algorithms to obtain keypoints and their corresponding descriptors. Translating feature descriptors is enough to enable matching. However, performance is drastically reduced in cross-feature detector cases, because current solutions assume common keypoints. This means that the same detector has to be used, which is rarely the case in practice when different descriptors are used. The low repeatability of keypoints, in addition to non-discriminatory and non-distinctive descriptors, make the identification of true correspondences extremely challenging. We present the first method tackling this problem, which performs feature descriptor augmentation targeting cross-detector feature matching, and then feature translation to a latent space. We show that our method significantly improves image matching and visual localization in the cross-feature scenario and evaluate the proposed method on several benchmarks.
\end{abstract}

%%%%%%%%% BODY TEXT

\section{Introduction}\label{sec:intro}
%\input{intro}
\begin{comment}
To add as motivation:
\begin{itemize}
    \item Investigation of cross-detector and tie to scenario
    \item Our method allows improvement over just translating descriptors
\end{itemize}    
\end{comment}

The present and future of mixed reality and autonomous devices entails a variety of devices operating and interacting in the same spaces. Complex applications require visual feedback, such as interaction of autonomous robots among themselves and the environment \cite{colslam} or shared augmented or virtual (AR/VR) experiences \cite{arcore, arkit, realitykit}. Usually, in these cases, devices require accurate positioning and access to recent and complete maps of the environment. 

New solutions envision providing localization and mapping as a cloud service \cite{location-based-services}, which has several advantages: enabling accurate perception to resource-constrained devices \cite{edgeslam}, building a more complete and up-to-date map by aggregating data from different sources, and providing information in the same global reference frame for all users.

%\begin{figure}[!htbp]
%    \includestandalone[width=\columnwidth]{diagrams/cover_matching}
%    \caption{Diagram showing our method. For any feature sets (e.g., $\mathcal{F}^{\text{SIFT}}$, and $\mathcal{F}^{\text{SuperPoint}}$) extracted with specific algorithms, our pipeline first augments the feature descriptors, then, augmented features are translated to a joint embedded space. After these two steps, two sparse feature sets extracted from different algorithms can be matched.}
%\end{figure}

\begin{figure}[!htbp]
        \centering
        %\begin{subfigure}[b]{\columnwidth}
        %    \centering           
        %    \includestandalone[width=\columnwidth]{diagrams/cover_matching_1}
        %    \caption[]{}
        %    \label{fig:cover-kps}
        %\end{subfigure}
        %\vskip\baselineskip
        %\includestandalone[width=\columnwidth]{diagrams/cover_matching_2}
            {
                    \usetikzlibrary{shapes,arrows}
                    \usetikzlibrary{positioning}
                    \usetikzlibrary{backgrounds,calc,positioning} 
                    \usetikzlibrary{intersections}
                    \newcommand*{\cemph}[2]{%
                        \tikz[baseline]\node[rectangle, fill=#1, rounded corners, inner sep=0.3mm, outer sep=0mm, anchor=base, minimum height=0cm, minimum width=0cm]{#2};%
                    }
                    \definecolor{blue-green}{rgb}{0.0, 0.87, 0.87}
                    \definecolor{limegreen}{rgb}{0.5529411764705883, 0.7803921568627451, 0.24313725490196078}
                    \begin{tikzpicture}[
                                        block/.style={
                                        draw,
                                        fill=white,
                                        rectangle,
                                        line width=1.2pt,
                                        font=\Large}
                                        ]

                        %\node[inner sep=0pt] (kptsplot) {\includegraphics[width=.8\textwidth]{images/homography_plots_crop/kpts_sift_sift-kornia_boosted_tr_embedded_superpoint-s3esti_superpoint_boosted_tr_embedded.pdf}};
                        %\node[inner sep=0pt, below=0cm of kptsplot] (amatches) {\includegraphics[width=.8\textwidth]{images/homography_plots_crop/kpts_matches_sift_sift-kornia_tr_embedded_baseline_superpoint-s3esti_superpoint_tr_embedded_baseline.pdf}};
                        %\node[inner sep=0pt, below=0cm of amatches.south] (bmatches) {\includegraphics[width=.8\textwidth]{images/homography_plots_crop/kpts_matches_sift_sift-kornia_boosted_tr_embedded_superpoint-s3esti_superpoint_boosted_tr_embedded.pdf}};
                        \node[inner sep=0pt] (bmatches) {\includegraphics[width=.5\columnwidth]{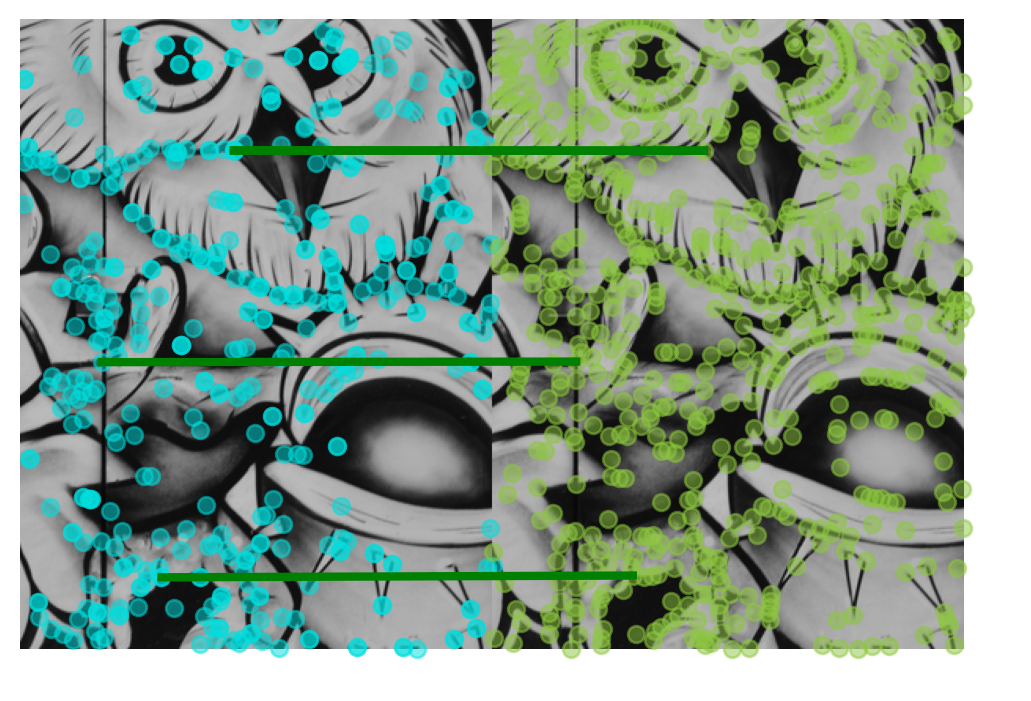}};
                        \node[inner sep=0pt, right=-0.3cm of bmatches] (amatches) {\includegraphics[width=.5\columnwidth]{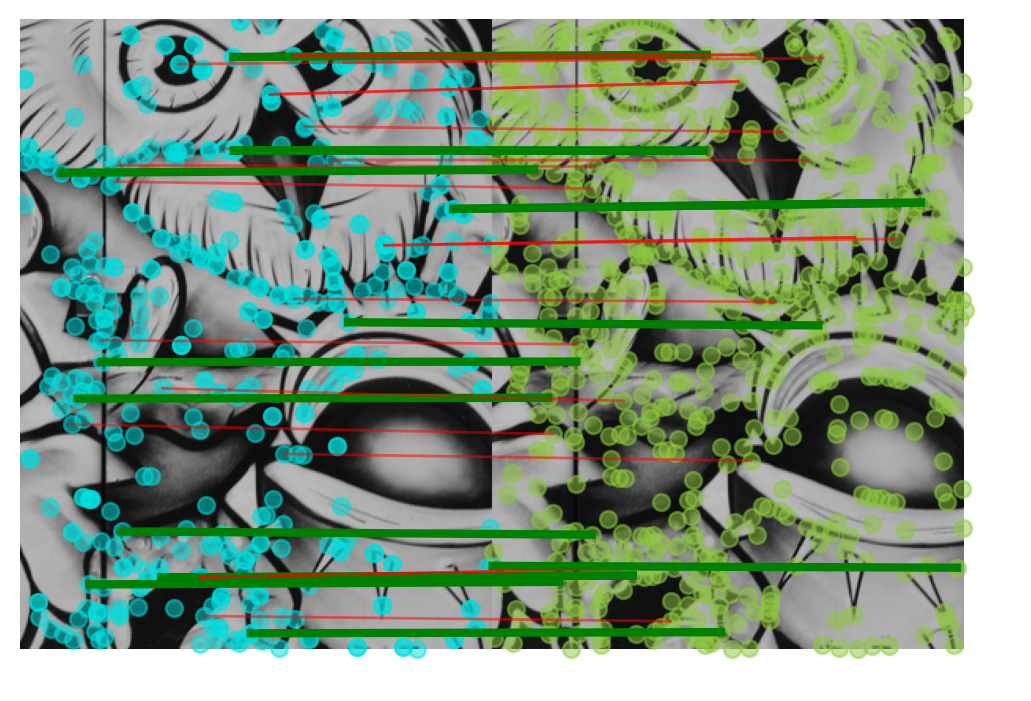}};
                        \node (leftn) at ($(bmatches.north east)$) {};
                        \node (rightn) at ($(amatches.north west)$) {};
                        \node (middlen) at ($(leftn)!0.5!(rightn)$) {};
                        \node[inner sep=0pt, above=-0.35cm of middlen] (kptsplot) {\includegraphics[width=.5\columnwidth]{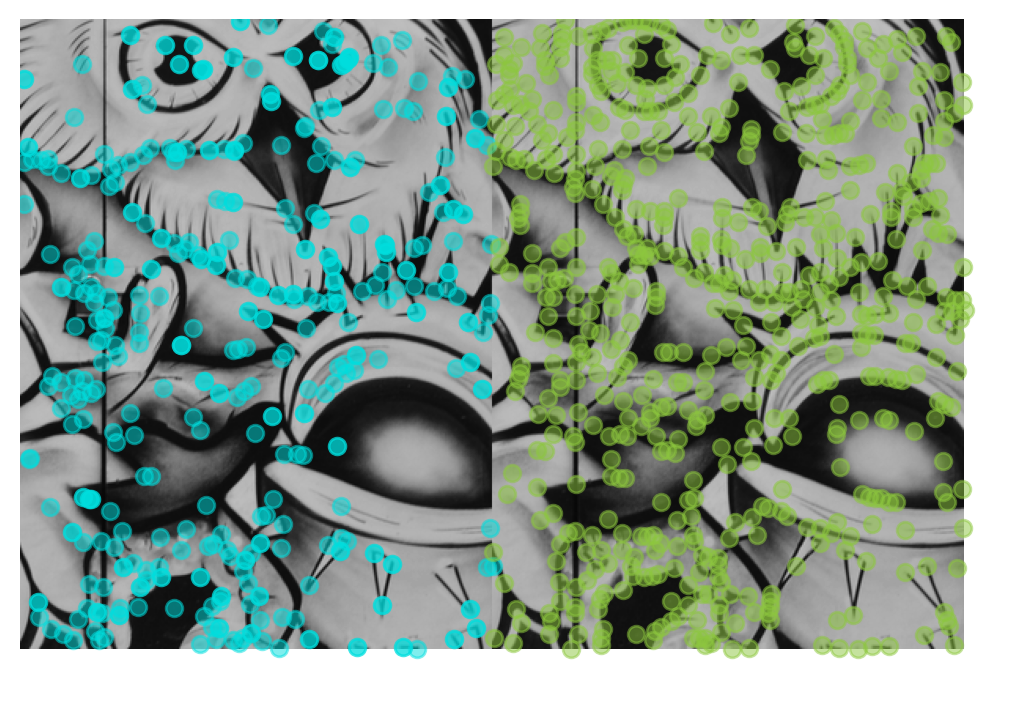}};
                        
                        \node (upnw) at ($(kptsplot.north west)+ (0.0,0.3)$) {};
                        \node (upn) at ($(kptsplot.north)+ (0.0,0.3)$) {};
                        \node (upne) at ($(kptsplot.north east)+ (0.0,0.3)$) {};
                        \node (labelsift) at ($(upnw)!0.5!(upn)$) {\cemph{blue-green}{$\mathcal{F}^{\text{SIFT}}$}};
                        \node (labelspt) at ($(upn)!0.5!(upne)$) {\cemph{limegreen}{$\mathcal{F}^{\text{SuperPoint}}$}};

                        \node (labelcrossdes) at ($(bmatches.south)+ (0.0,-0.1)$) {\cite{crossdes}};
                        \node (labelcrossdes) at ($(amatches.south)+ (0.0,-0.1)$) {Ours};
                        %\node (table) at ($(amatches.south west)+ (-0.8,-1.4)$) {\begin{tabularx}{\columnwidth}{|>{\centering\arraybackslash}X|>{\centering\arraybackslash}X|}\hline \xmark & \cmark \\ \hline \cmark & \cmark  \\\hline \end{tabularx}};
                        %\node (tablelabels) at ($(table.west)+ (-0.0,0.0)$) {\begin{tabularx}{0.3\columnwidth}{|>{\centering\arraybackslash\hsize=.45\hsize}X|}\hline Any detector \\ \hline Any descriptor  \\\hline \end{tabularx}};
                        %\node (table) at ($(amatches.south west)+ (0.0,0.0)$) {\begin{tabularx}{0.8\columnwidth}{|>{\centering\arraybackslash}X|>{\centering\arraybackslash}X|>{\centering\arraybackslash}X|}\hline \xmark & \cmark \\ \hline \cmark & \cmark  \\\hline \end{tabularx}};

                        %\node[align=center] (a) at ($(amatches.north west)+ (-1.0,0.0)$){\begin{tabularx}{0.3\columnwidth}{|*{3}{X|}}\hline  Method & Any detector & Any descriptor \\ \hline \cite{crossdes} & \xmark & \cmark\\ \hline Ours & \cmark & \cmark \end{tabularx}};
                        %\node[align=center, below=0.5cm of amatches] (b) {\begin{tabularx}{\columnwidth}{|*{3}{X|}}\hline  Method & Any detector & Any descriptor \\ \hline Ours & \cmark & \cmark\\ \end{tabularx}};
                    \end{tikzpicture}
            }
        \caption[]{\small \textbf{Top:} difference in keypoint distribution for $\text{SIFT}$ and $\text{SuperPoint}$ features in the \textbf{same} image from HPatches-sequences \cite{hpatches}. \textbf{Bottom:} Inlier and outlier matches for \cite{crossdes} and our method. Even matching cross-algorithm features between the same image is challenging for the state-of-the-art. Our method is explicitly designed for cross-algorithm feature matching scenarios. Matching performed with nearest neighbor retrieval, ratio test and RANSAC.} 
        \label{fig:cover-hpatches}
\end{figure}
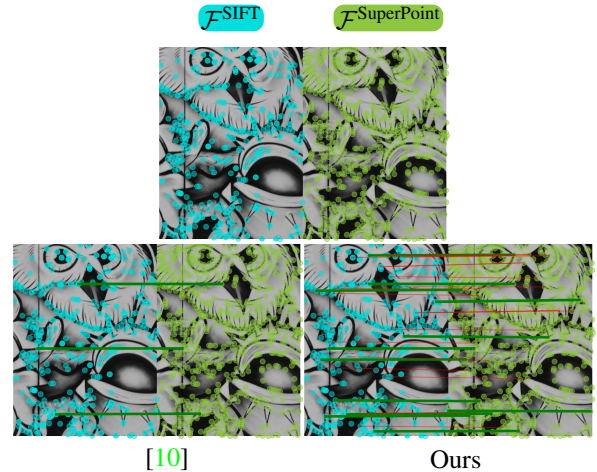

Most modern localization and mapping systems for mixed reality and robotics rely on visual data \cite{hloc,inloc,back2feature}. The possibility of streaming full images to the cloud is however limited by bandwidth availability in the network and privacy reasons. To address this, recent works \cite{crossdes,gomatch} propose that  devices extract and share their local sparse features to access cloud localization services.

A feature extraction algorithm, can be understood in two stages: a feature detector, typically followed by a feature descriptor. Handcrafted methods, e.g., \cite{sift,orb} adhere to a two stage detect then describe, while learning based methods \cite{superpoint, d2net} perform detect and describe, i.e., compute keypoints and descriptors, at the same time. More recently, a decoupled method has been presented in \cite{dedode}, which leverages dedicated networks for each task. Depending on the application, e.g., visual localization, SLAM, odometry or mapping, some feature extraction algorithms might be favoured over others, based on their computational cost, size (with regards to scalability), descriptiveness and/or robustness. Current multi-agent methods that involve sharing sparse features between devices require the same algorithm, to be used \cite{maplabv2}. However, this severely hinders large scale device interoperability and collaboration among heterogeneous smart platforms.

In previous work~\cite{crossdes} it was demonstrated that cross-descriptor matching (matching across different descriptors), $\text{D}$, such as SIFT~\cite{sift} and HardNet~\cite{hardnet} is possible by translating to a learnt joint embedding space. The results are a first step towards collaborative mapping and cross-algorithm localization scenarios. However, it is assumed that the same detector is used for all descriptors, an unrealistic setting when using different devices or algorithms.

In this work, we provide three contributions towards enabling cross-algorithm (cross-detector and cross-descriptor) feature matching.
The first contribution of our paper is to remove the assumption of a common detector held in \cite{crossdes}, which was an initial simplification of cross-algorithm scenario which does not hold in practice. We demonstrate that in true cross-algorithm scenarios it is not sufficient to simply translate descriptors, $\text{D}$, for accurate localization. When using different detectors for matching images, the number of repeated points drops across different views, which significantly reduces performance, as can be seen in our results in \cref{fig:hpatcheseval} and \cref{tab:aachen}.
As our second contribution, and addressing the problem encountered in the first contribution, we propose a detector and descriptor-specific feature augmentation pipeline that tackles matching in cross-detector scenarios, coupled with translation networks, which together enable full cross-algorithm matching.
Finally, our last contribution is evaluating the proposed method on the image matching task with the HPatches benchmark \cite{hpatches}, and on the visual localization task in the Aachen Day and Night v1.1 \cite{aachen} and 7Scenes~\cite{shotton2013scene} benchmarks. Our results show that by augmenting the features with detector and descriptor-specific pipelines, we can significantly improve localization accuracy in cross-algorithm scenarios.

\section{Related work}\label{sec:relatedwork}

\subsection{Sparse feature representations}
Sparse features are the combination of a detected interest point, or keypoint, and its corresponding descriptor. Features, traditionally extracted with handcrafted methods and more recently learning-based, are intended to be repeatable and discriminative. Repeatability referring to how robustly a 3D landmark can be identified as an interest point in 2D across changes in the scene due to viewpoint or other geometric transforms, illumination, and image noise \cite{scalenaffine-repeatability,bias-repeatability}. Discriminability meaning that an interest point is distinctive and can be clearly distinguished among others\cite{learning-affine-discriminability}.

The most well-known methods are the ones that over time have proven to be the most reliable in these two metrics. Handcrafted detectors are often blob-like \cite{sift} or corner-like \cite{harriscorners}. Some detect points of interest plus scale and orientation of the feature \cite{sift} which gives a better understanding of the support region of the point; some detectors extract affine features \cite{asift} that can be more reliable given viewpoint changes of planar surfaces. 
More recently, learnt detectors and/or descriptors have been designed to perform well in these evaluation metrics \cite{r2d2,learning-affine-discriminability, alike} or directly for feature matching \cite{superpoint}.

Both handcrafted and learnt detectors and descriptors have different advantages depending on the application. For SLAM, lightweight methods that can run in real-time are preferred, with detectors like FAST \cite{fast} and descriptors such as BRIEF \cite{brief}. For Structure-from-Motion (SfM) higher number of more descriptive features will provide more accurate results, like SIFT \cite{sift} or SuperPoint \cite{superpoint}. Since in this work we focus on visual localization in SfM maps, we will evaluate our method in the latter set of features.

\subsection{Feature matching and localization} 
Visual localization has the aim of estimating a rotation and translation up to scale for an image given a set of reference images. In this case, we explore feature matching techniques for sparse feature representations. 

Traditional local feature matching techniques rely on nearest neighbor retrieval in the descriptor space to find candidate matches. Candidate matches are then filtered using heuristics such as mutual check and Lowe's ratio test \cite{sift}, relying on the discriminability of the descriptors. Then a pose is estimated for these 2D-2D or 2D-3D correspondences, together with an outlier rejection technique like RANSAC \cite{ransac}. 

Learning-based methods have recently gained popularity as substitutes to the traditional pairwise correspondence estimation \cite{superglue,lightglue}. SuperGlue~\cite{superglue} presented a first step which aggregates context of features in two covisible images to enhance them based on self and cross-attention, and a second step estimating correspondences with an assignment matrix. Similarly \cite{omniglue} follows an approach including knowledge from a foundation model, which predicts where the edges should be placed in the feature graph instead of using a fully-connected architecture.

GoMatch \cite{gomatch} and DGC-GNN \cite{dgcgnn} argue that visual descriptors, which are storage-demanding but include rich information synthesised from the pixel information from the support region of every keypoint, can be discarded. Instead, the geometry of the sparse features only is enough for localization. Geometric information is augmented with self and cross attention into a geometric descriptor, used for matching. DGC-GNN has shown great improvement over GoMatch, however, utilizing feature descriptors is still more accurate, and it is not clear whether a geometric-only method could handle cross-detector cases given low feature repeatability and lack of more descriptive information.

Feature matching can align covisible images in a local context, but in larger scales more complex frameworks are needed, which sometimes use a hierarchical approach \cite{hloc}.  In these approaches, Structure-from-Motion (SfM) is used to build a map from covisible images, obtained by matching global descriptors as NetVLAD \cite{netvlad}. This process can be repeated to find covisible pairs between queries and map/reference images.

\subsection{Augmenting features}
The concept behind feature augmentation is improving descriptors distinctiveness for matching. This research has mostly encompassed context aggregation from geometrical and visual information contained in other keypoints. Some works used CNNs \cite{contextdesc} to augment local descriptors with context aggregated from the whole image. However this method requires access to the full image to incorporate more visual context. Working solely on the feature space, SConE \cite{scone} presents a siamese network to encode extra geometric information from adjacent keypoints. SuperGlue \cite{superglue} introduced a first attention-based approach aimed at enhancing each feature descriptor by aggregating context from all the other features in the image (self-attention) as well as in the covisible reference image (cross-attention), followed by a learnt correspondence estimation step. This first step proved effective in challenging conditions with many similarities or little visual overlap. Since, works like FeatureBooster \cite{featurebooster} focus solely on the feature augmentation part.

\section{Problem setting}\label{sec:problem_setting}

Expanding from \cite{crossdes}, we envision a set of camera-equipped devices working in the same environment, each device using a certain local feature extraction algorithm $\text{A}^{\alpha,a} = (\text{E}^\alpha, \text{D}^a)$ with a certain description method $\text{D}^a \in \mathcal{D}$ and in our case also a different keypoint detector $\text{E}^\alpha \in \mathcal{E}$. Given detector and descriptor heterogeneity, a trivial solution would be to re-extract features for every new image given a main local feature extraction algorithm, however, this involves devices streaming full images at high rates which may not be possible due to bandwidth limitations. We instead tackle the cross-algorithm matching problem: using a different detector (cross-detector) and descriptor (cross-descriptor) for query and reference images, with the limitation that devices only share sparse features. Our strategy relies on using context from adjacent features to augment each feature, making the repeatable feature set easier to identify and subsequently match. After, a translation step makes descriptors comparable.

\section{Method}\label{sec:method}
\begin{figure}[!htbp]
    \centering
    %\includestandalone[width=0.8\columnwidth]{diagrams/method}
    \includegraphics[width=0.8\columnwidth]{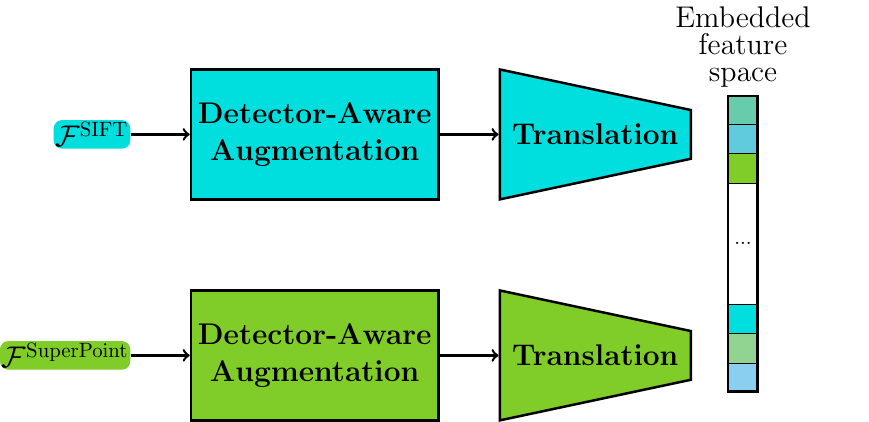}
    \caption{Diagram showing an execution of our method. For any feature sets (e.g., $\mathcal{F}^{\text{SIFT}}$, and $\mathcal{F}^{\text{SuperPoint}}$) extracted with specific algorithms, our pipeline first augments the feature descriptors, then, augmented features are translated to a joint embedded space. After these two steps, two sparse feature sets extracted from different algorithms can be matched.}
    \label{fig:method}
\end{figure}

To solve the cross-algorithm matching problem, we design a first step, named \textbf{detector-aware descriptor augmentation}, that enhances descriptors for optimizing cross-detector matching. Then, a second step, referred as \textbf{cross-algorithm feature translation}, encodes heterogeneous descriptors to the same latent space, where they can be matched.
An overview of the method is presented in \cref{fig:method}.

\subsection{Detector-aware descriptor augmentation}\label{ssec:dada}
The first part of our solution consists of a feature augmentation step that enhances the descriptors for cross-detector matching. This is achieved by designing each branch in the architecture for a specific detector-descriptor pair $(a, \alpha)$, and training together networks corresponding to the descriptor which is to be augmented. Each branch follows the architecture presented in FeatureBooster \cite{featurebooster}.

\begin{figure}[!htbp]
    \centering
    %\includegraphics[width=\textwidth]{images/methodtrain:val.png}
    %\includestandalone[width=0.55\columnwidth]{diagrams/training-overview}
    \includegraphics[width=0.55\columnwidth]{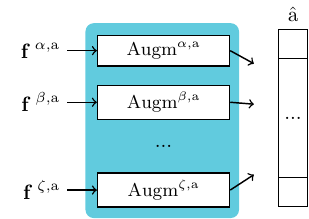}
    %\caption{Detailed overview of our method. As in our results, we show the architecture for DoG and SuperPoint as detectors and SIFT and SuperPoint as descriptors. At inference time, we use a \textbf{detector-aware descriptor augmentation module} trained on a particular algorithm (detector-descriptor pair) $\{\text{E},\text{D}\}$, as well as \textbf{feature encoders and decoders for feature translation} trained for the augmented feature spaces. At training time, the \textbf{detector-aware descriptor augmentation} networks are trained together cross-detector for the same descriptor, improving cross-detector matching in the augmented space. This means before the translation step, features are projected to a space that enhances them for cross-detector feature matching.}
    \caption{Architecture for training the contribution detailed in \cref{ssec:dada} and summarized in \cref{eq:augm}. Detector-aware feature augmentation networks are detector and descriptor specific, and are trained together for the same descriptor algorithm given a set of feature detector algorithms to maximize the Cross-Detector Average Precision (\cref{eq:cdap}), computed for the augmented feature space.}
    \label{fig:trainingaugm}
\end{figure}

%For features described with descriptor algorithm $\text{D}^{a}$, features are augmented depending on feature detection algorithm $\text{E}^{\alpha}$. This yields algorithm-specific feature augmentation branches as can be seen in \cref{fig:detailedmethod}.
%Each feature augmentation branch is devised for a specific detector-descritor pair $(a, \alpha)$, where the descriptor $\text{D}^{a}$ is enhanced for the specific keypoints detected from $\text{E}^\alpha$
%This way, branches for the descriptor $\text{D}^{a}$ are designed in parallel to enhance descriptors given a set of feature detectors $\mathcal{E}$. Each separate branch follows the architecture presented in FeatureBooster \cite{featurebooster}.

%Device running algorithm $\text{A}^{ij}$ for feature extraction and description produces a set of features $\mathcal{F}^{ij} = \{f^{ij}\subb{1}, \ f^{ij}\subb{2}, \ \dots \}$.  Each feature $f^{ij}_k$ consists of a keypoint and a descriptor, $f^{ij}_k = ( \textbf{k}^i_k, \textbf{d}^j_k )$, $\textbf{k}^i_k \in \mathbb{R}^5$, $\textbf{d}^j_k \in \mathbb{R}^{n^j}$. The keypoint $\textbf{k}^i_k = (x, y, s, \theta, c)^i_k$ represents geometric information: 2D coordinates in the image plane, scale, orientation and confidence score respectively. $n^j$ is the descriptor dimension for description algorithm $\text{D}^j$. 

A device running algorithm $\text{A}^{\alpha,a}$ for feature detection and description produces a set of features $\mathcal{F}^{\alpha,a} = \{\mathbf{f}^{\alpha,a}\subb{1}, \ \mathbf{f}^{\alpha,a}\subb{2}, \ \dots \}$ per image.  Each feature $\mathbf{f}^{\alpha,a}_i = ( \textbf{k}^\alpha_i, \textbf{d}^a_i )$ consists of a keypoint $\textbf{k}^\alpha_i \in \mathbb{R}^5$ and a descriptor $\textbf{d}^a_i \in \mathbb{R}^{n^a}$, where $\textbf{k}^\alpha_i = (x, y, s, \theta, c)^\alpha_i$ represents geometric information: 2D coordinates in the image plane, scale, orientation and confidence score, respectively; and $n^a$ is the descriptor dimension for description algorithm $\text{D}^a$.

First, the branch encodes geometric information into every descriptor. The geometric information contained in each keypoint $\textbf{k}^\alpha_i$ is projected into the corresponding descriptor $\textbf{d}^a_i$ with an MLP (Multi-layer perceptron). The descriptor is also projected with another MLP, and both quantities are added element-wise:
\begin{equation}\label{eq:augmmlp}
    \textbf{d}^{a}_{i, \textit{enc}}
    \leftarrow \text{MLP}^{\alpha,a}\subb{keypt}(\textbf{k}^\alpha_i) + \text{MLP}^{\alpha,a}\subb{desc}(\textbf{d}^a_i).
\end{equation}

This yields a set of encoded descriptors %$\{ {\textbf{d}^a\subb{1}}^{\textit{enc}}, {\textbf{d}^a\subb{2}}^{\textit{enc}}, \dots \}$ 
$\{ \textbf{d}^a\subb{1, \textit{enc}}, \textbf{d}^a\subb{2, \textit{enc}}, \dots \}$ 
which are projected with a transformer network to aggregate context between them:
\begin{equation}\label{eq:augmT}
    \{ {\textbf{d}^{\hat{a}}\subb{1}}, {\textbf{d}^{\hat{a}}\subb{2}}, \dots \} \leftarrow \text{T}^{\alpha,a}(\{ \textbf{d}^a\subb{1, \textit{enc}}, \textbf{d}^a\subb{2, \textit{enc}}, \dots \}).
\end{equation}
And simply aggregating \cref{eq:augmmlp} and \cref{eq:augmT} together:
\begin{equation}\label{eq:augm}
    \{ {\textbf{d}^{\hat{a}}\subb{1}}, {\textbf{d}^{\hat{a}}\subb{2}}, \dots \} \leftarrow {\text{Augm}}^{\alpha,a}(\{ (\textbf{k}^\alpha\subb{1}, \textbf{d}^a\subb{1}), (\textbf{k}^\alpha\subb{2}, \textbf{d}^a\subb{2}), \dots \}).
\end{equation}

The transformer network $\text{T}^{\alpha,a}$ treats features as nodes in a fully-connected graph and then applies an attention mechanism recurrently to aggregate context from all nodes. The result is a new descriptor for each keypoint that is augmented given all the other features detected. The attention mechanism is computationally heavy, therefore we employ an Atention-Free Transformer \cite{attention-free-transformer} for this context-aggregation step, as in FeatureBooster \cite{featurebooster}.

The output of the branch is a new set of augmented features $\mathcal{F}^{\alpha,\hat{a}} = \{ \mathbf{f}\subb{1}^{\alpha,\hat{a}}, \mathbf{f}\subb{2}^{{\alpha,\hat{a}}}, \dots \}$, with $\mathbf{f}^{{\alpha,\hat{a}}}_i = ( \textbf{k}^\alpha_i, \textbf{d}^{\hat{a}}_i )$, $\textbf{d}^{\hat{a}}_i \in \mathbb{R}^{n^a}$ (the original geometric information of the feature is kept and the descriptor is augmented). As illustrated in \cref{fig:trainingaugm}, these two steps are detector and descriptor-specific.

\paragraph{Detector-aware descriptor augmentation loss:}
The loss is defined to maximize Average Precision (AP) \cite{auc} in matching with nearest neighbors. Given the feature description algorithm $\text{D}^a$ and a set of feature detection algorithms $\mathcal{E}$, we define the cross-detector augmentation loss as:

\begin{equation}
    \mathcal{L}\subb{cd} = \sum_{\alpha}^{|\mathcal{E}|} (1-\frac{1}{K}\sum_i^K \text{CDAP}(\mathbf{f}^{{\alpha,\hat{a}}}_i) ), 
\end{equation}
where $K$ is the size of the augmented feature set ${|\mathcal{F}^{\alpha,\hat{a}}| = K}$. We define the Cross-Detector Average Precision ($\text{CDAP}$) as:

\begin{equation}\label{eq:cdap}
    \text{CDAP}(\mathbf{f}^{{\alpha,\hat{a}}}_i) = \frac{1}{|\mathcal{E}|} \sum_{\beta}^{|\mathcal{E}|} \text{AP}(\mathbf{f}^{{\alpha,\hat{a}}}_i, \mathcal{F}^{\beta, \hat{a}})
\end{equation}
and where the AP is defined as in \cite{featurebooster, auc}. To compute the AP, a differentiable approximation called FastAP is used. More details can be found in \cite{featurebooster, fastap}. We add a boosting loss that ensures the output descriptors perform better than the original descriptors for the given matching objective:
\begin{equation}
    \mathcal{L}\subb{boost} =  \frac{1}{K} \sum_{\alpha}^{|\mathcal{E}|} \sum_i^K \max(0, \frac{\text{CDAP}(\mathbf{f}^{{\alpha,\hat{a}}}_i)}{\text{CDAP}(\mathbf{f}^{{\alpha,a}}_i) } - 1)
\end{equation}
and the total loss is a weighted sum of both:
\begin{equation}
    \mathcal{L}\subb{augmentation} = \mathcal{L}\subb{cd} + \lambda \mathcal{L}\subb{boost}.
\end{equation}

A diagram of the architecture of this contribution at training time can be seen in \cref{fig:trainingaugm}.

\subsection{Cross-algorithm feature translation}\label{ssec:translation}
The second part of our solution aims at projecting augmented features to a joint embedded space, inspired by \cite{crossdes}. The translation is detector-agnostic. 

For a given keypoint $\textbf{k}_k$ and given a pair of feature description algorithms $\text{D}^a, \text{D}^b \in \mathcal{D}$. The corresponding descriptors for this same keypoint are augmented into descriptors $\textbf{d}_k^{\hat{a}}, \textbf{d}_k^{\hat{b}}$ as explained in the previous section, with $|\mathcal{F}^{\hat{a}}| = |\mathcal{F}^{\hat{b}}| = \ldots = K$. We train an encoder and a decoder for each descriptor algorithm.
These networks, $(\text{Enc}^{\hat{a}}, \text{Dec}^{\hat{a}}$) and $(\text{Enc}^{\hat{b}}, \text{Dec}^{\hat{b}})$, can be used to translate the descriptor vectors to a latent space \textit{EMB} and back:
\begin{align}
    \textbf{d}_k^{\text{EMB}} &= \text{Enc}^{\hat{a}}(\textbf{d}_k^{\hat{a}})\\
    \textbf{d}_k^{\hat{a}} &= \text{Dec}^{\hat{a}}(\textbf{d}_k^{\text{EMB}})
\end{align}
with $\textbf{d}_k^{\text{EMB}} \in \mathbb{R}^{n^{\text{EMB}}}$. Encoders and decoders can be chained to translate in between augmented feature spaces:
\begin{equation}\label{eq:direct-translation}
    \textbf{d}_k^{\hat{b}} = \text{Dec}^{\hat{b}} ( \text{Enc}^{\hat{a}} (\textbf{d}_k^{\hat{a}}) ) = \text{P}^{\hat{a}\rightarrow \hat{b}} (\textbf{d}_k^{\hat{a}})
\end{equation}

This operation applies to all descriptor algorithms in $\mathcal{D}$. 

\paragraph{Feature translation loss:} Following \cite{crossdes}, these encoders and decoders are trained on translation and matching losses.

A direct translation loss ensures that the translation in between augmented descriptor spaces is as accurate as possible. Given two augmented descriptors for the same keypoint,$\textbf{d}_k^{\hat{a}}$, $\textbf{d}_k^{\hat{b}}$, the direct translation loss is defined as:
\begin{equation}
\mathcal{L}\subb{dirtr} = \frac{1}{|\mathcal{D}|^2} \sum_{\hat{a}}\supp{|\mathcal{D}|} \sum_{\hat{b}}\supp{|\mathcal{D}|} \frac{1}{K} \sum_k\supp{\text{K}} \| \text{P}^{\hat{a}\rightarrow \hat{b}} (\textbf{d}_k^{\hat{a}}) - \textbf{d}_k^{\hat{b}} \|
\end{equation}

The matching loss ensures that descriptors in the embedded space are discriminative, ensuring the embedded space can be used for matching. The triplet margin loss is used:
\begin{align}
\mathcal{L}\subb{match}^{(\hat{a}, \hat{b})} &= \\
&\frac{1}{K} \sum_k\supp{K} \max(0, m + \text{pos}(\textbf{d}_k^{\hat{a}}, \textbf{d}_k^{\hat{b}}) - \text{neg}(\textbf{d}_k^{\hat{a}}, \mathcal{F}^{\hat{b}}))\nonumber
\end{align}
where $\text{pos}(\textbf{d}_k^{\hat{a}}, \textbf{d}_k^{\hat{b}})$ is the distance from $\textbf{d}_k^{\hat{a}}$ to the positive sample $\textbf{d}_k^{\hat{b}}$ in the embedded space. The negative distance $\text{neg}(\textbf{d}_k^{\hat{a}}, \mathcal{F}^{\hat{b}})$ is the distance in the embedded space to the closest negative sample given the whole set of features $\mathcal{F}^{\hat{b}}$. Variable $m$ is used to enforce some margin in the embedded space. More details can be found in \cite{crossdes}. The total matching loss is the sum for all descriptor algorithms:

\begin{equation}
    \mathcal{L}\subb{match} = \frac{1}{|\mathcal{D}|^2} \sum_{\hat{a}}^{|\mathcal{D}|} \sum_{\hat{b}}^{|\mathcal{D}|} \mathcal{L}\subb{match}^{(\hat{a},\hat{b})},
\end{equation}
and the final loss is the weighted sum:
\begin{equation}
    \mathcal{L}\subb{translation} = \mathcal{L}\subb{dirtr} + \gamma \mathcal{L}\subb{match}.
\end{equation}

\section{Implementation and evaluation}\label{sec:expsetup}

%\subsection{Feature detectors and descriptors:}
The features used for this evaluation are SIFT \cite{sift} and SuperPoint \cite{superpoint} chosen given their prevalence and performance. Detection of SIFT features is performed with the Difference of Gaussians (DoG) \cite{sift} implementation from COLMAP \cite{colmap}, while description is performed with the KORNIA library \cite{kornia}, with size $n^{\text{SIFT}} = 128$. For SuperPoint, the model and weights released with SuperGlue \cite{superglue} are used, with size $n^{\text{SuperPoint}} = 256$. The geometric properties of scale and orientation are not computed for SuperPoint, hence we couple it with the scale and orientation estimator S3-Esti \cite{s3esti}. We implement all networks with PyTorch \cite{pytorch} and use ReLU activation functions after each linear layer except for the last one.

\subsection{Detector-aware descriptor augmentation}
\paragraph{Network architecture:} The MLPs for geometric encoding have output size $(32, 64, 128, n^a, n^a)$ where $n^a$ is the size of the descriptor. The MLP for descriptor encoding has output sizes $(256, n^a)$. The transformer network has $4$ attention layers for SIFT and $9$ for SuperPoint.

\paragraph{Training methodology:} The MegaDepth dataset \cite{megadepth} is used for training all the networks. The same train and validation scenes as in \cite{featurebooster} are selected. Similarly to \cite{featurebooster, d2net}, $500$ pairs of images with an overlap score (defined in \cite{d2net}) in between $[0.1,1]$ are sampled for each scene. Each image in the pair is detected and described with all combinations of DoG, SuperPoint as detectors and SIFT, SuperPoint as descriptors. Networks for the same descriptor are trained together for $2^n$ combinations of detector to detector matching, with $n$ being the number of detection algorithms $|E|$, in our case $2$. The 2D coordinates of keypoints are normalized given image width and height, scale is normalized to be in $[-0.5, 5]$, orientation is expressed in radians and normalized to be in $(-\pi, \pi]$, and scores are normalized within features in an image. Ground truth matches are found for points with reprojection error lower than $3$ pixels. 
Networks for the same descriptor are trained together as detailed in \cref{fig:trainingaugm}, with early stopping for a total of $23$ epochs for SIFT and $25$ epochs for SuperPoint descriptors. We use AdamW optimizer \cite{adamw}, batch size of $16$, $\lambda = 10$, learning rate increasing linearly until \num{1e-4} for the first \num{500} iterations and cosine decay after. We train SIFT augmentation with detectors $(\alpha, \beta) = (\text{DoG}, \text{SuperPoint})$, and SuperPoint augmentation with $(\alpha, \beta) = (\text{SuperPoint}, \text{DoG})$.

\subsection{Cross-algorithm feature translation}\label{subs:caftranslation}
\paragraph{Network architecture:}  The encoders and decoders have layers with output sizes $[256, 256, n^{\text{EMB}}]$for SuperPoint and $[1024,1024, n^{\text{EMB}}]$ for SIFT, with $n^{\text{EMB}} = 256$. Batch-normalization is used after each layer except the last one. The outputs of the networks are $L_2$ normalized.

\paragraph{Training methodology:} The training dataset is the Oxford-Paris revisited \cite{oxfordparisrev}, where \num{12.5e6} points are extracted for DoG and SuperPoint, and \num{19e6} points are extracted from DoG and Superpoint combined. When training, positive samples are all other descriptors extracted for the exact same keypoint, whereas negative samples are found for any keypoint different than the original. As in \cite{crossdes}, we train with Adam optimizer \cite{adam}, with learning rate \num{1e-3}, $\gamma = 0.1$ and $m=1$. A batch size of $4096$ is used. Encoders and decoders are trained for $17$ epochs.

\subsection{Experiment setup}\label{subs:expsetup}
The proposed method is evaluated in the task of image matching in the HPatches dataset \cite{hpatches}, and visual localization in the Aachen Day and Night dataset v1.1 \cite{aachen} and 7Scenes dataset~\cite{shotton2013scene}. These datasets are chosen because of their relevance for relative pose estimation in different conditions, environments and at different scales. 

We compare our work against \cite{crossdes}, which represents the current state-of-the-art. This baseline was re-trained given the instructions provided by the authors, for DoG points extracted in the Oxford-Paris revisited \cite{oxfordparisrev} for descriptors SIFT-kornia and Superpoint. The network architecture is the same as detailed in \cref{subs:caftranslation} and the size of the embedded space $n^{\text{EMB}}$ is set to 128. Networks are trained with \num{6.5e6} patches extracted with DoG, for $5$ epochs. We evaluate our method and the baseline for direct translation of the query descriptor to the map descriptor, which entails using the query encoder to obtain the query features in the embedded space and then the map decoder to project back to the map descriptor space. We also obtain results for matching directly in the embedded space.

Evaluations are classified in \emph{Homogeneous} and \emph{Heterogeneous} cases. In homogeneous cases, the map and the query use the same descriptor, therefore no translation step is needed and features can be matched directly. These cases are labeled as \emph{$\text{Direct}$}. We report results for the augmentation step of our method only, as \cref{eq:augm} shows, and we label these results as \emph{$\text{Ours}^{A-}$}. We also report results translating to embedded space for our full method (\emph{$\text{Ours}^{\text{EMB}}$}) as well as the baseline \cite{crossdes} (\emph{$\text{C-D}^{EMB}$}). In heterogeneous cases, only the baseline and our work can be evaluated, since descriptors between query and map become incompatible for matching. We include \emph{$\text{Ours}^{\text{EMB}}$} and \emph{$\text{C-D}^{EMB}$} as defined before, but also \emph{$\text{Ours}$} and \emph{$\text{C-D}$} which in this case always refers to translating from query algorithm to map algorithm, using direct translation as in \cref{eq:direct-translation}. In our case, we use the augmentation module for the query features as well.

\paragraph{Image matching:} For the task of image matching, we evaluate our method, trained on SIFT and SuperPoint, in the HPatches sequences dataset \cite{hpatches}. This dataset contains $116$ sequences of $6$ images each, and for each sequence, either illumination or viewpoint is varied. Ground truth homographies are provided. We evaluate Mean Matching Accuracy (MMA) \cite{performanceevallocaldes} and inlier number for all sequences except eight that are excluded as in \cite{d2net}. This analysis is done for varying thresholds of reprojection error, from $1$ to $10$ pixels. MMA corresponds to the ratio of correctly matched points, and reveals how  accurate the method can be. Measuring the number of inliers, in our case, shows how many repeated keypoints can be correctly identified after matching. 

We follow the \emph{Homogeneous} - \emph{Heterogeneous} separation aforementioned. For homogeneous cases, MMA and inlier number are evaluated given the same feature extraction algorithm on all images of the sequence. For heterogeneous cases the first image (treated as a query) has a different feature extraction algorithm than the rest of the sequence (the map). In these cases, the labels indicate the configuration for $map \ - \ query$ algorithms. Query to map translations and translations to the embedded space are performed.

\paragraph{Visual localization:}\label{parag:expsetup_vl} We evaluate our method for visual localization with features SIFT and SuperPoint on two datasets: the Aachen Day and Night v1.1 dataset \cite{aachen}, captured outdoors and containing $6697$ map images and $824$ daytime and $191$ nighttime query images; and 7Scenes~\cite{shotton2013scene}, depicting 7 indoor scenes with a total of 46 sequences from which 18 were used for testing, as in the original split. HLoc\cite{hloc} is used for map creation and query localization. The map is built by matching each map image to its 20 closest spatial neighbors for Aachen. For 7Scenes, we employed the models provided by PixLoc~\cite{back2feature}. The same feature type is used to create the map. To localize query images, for Aachen, NetVLAD \cite{netvlad} is employed as a global descriptor to find the 50 pairs with the highest covisibility for each query, and the top-10 retrievals from DenseVLAD~\cite{torii201524} provided by the PixLoc authors were used in 7Scenes. For local feature matching, a nearest-neighbor approach with mutual check is employed, except when using SuperGlue~\cite{superglue} ($^\text{SG}$) or LightGlue~\cite{lightglue} ($^\text{LG}$). The error for the estimated pose of the queries with respect to the map was evaluated differently in each case: in the Aachen dataset we evaluated it in terms of percentage of localized queries within certain error bounds, while in 7Scenes we evaluate the median errors along testing sequences. We follow the \emph{Homogeneous} - \emph{Heterogeneous} separation aforementioned, and include FeatureBooster \cite{featurebooster} (\emph{FB}) for the homogeneous cases.

\section{Results}
\subsection{Image matching}
    \begin{figure*}[!htbp]
        \centering
        \begin{subfigure}[b]{0.85\textwidth}
            \centering           \includegraphics[width=\textwidth]{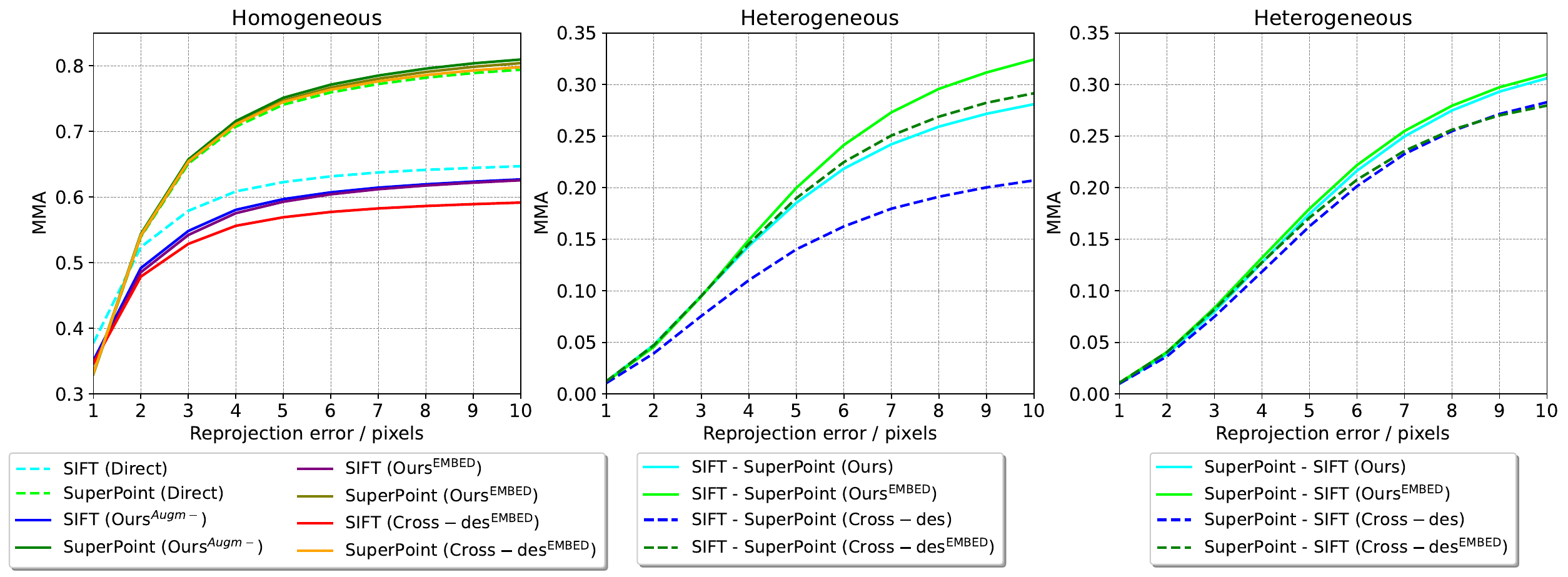}
            \caption[]{{\small Results for MMA in the HPatches sequences dataset}}    
            \label{fig:mma}
        \end{subfigure}
        \vskip\baselineskip
        \begin{subfigure}[b]{\textwidth}   
            \centering 
            \includegraphics[width=0.85\textwidth]{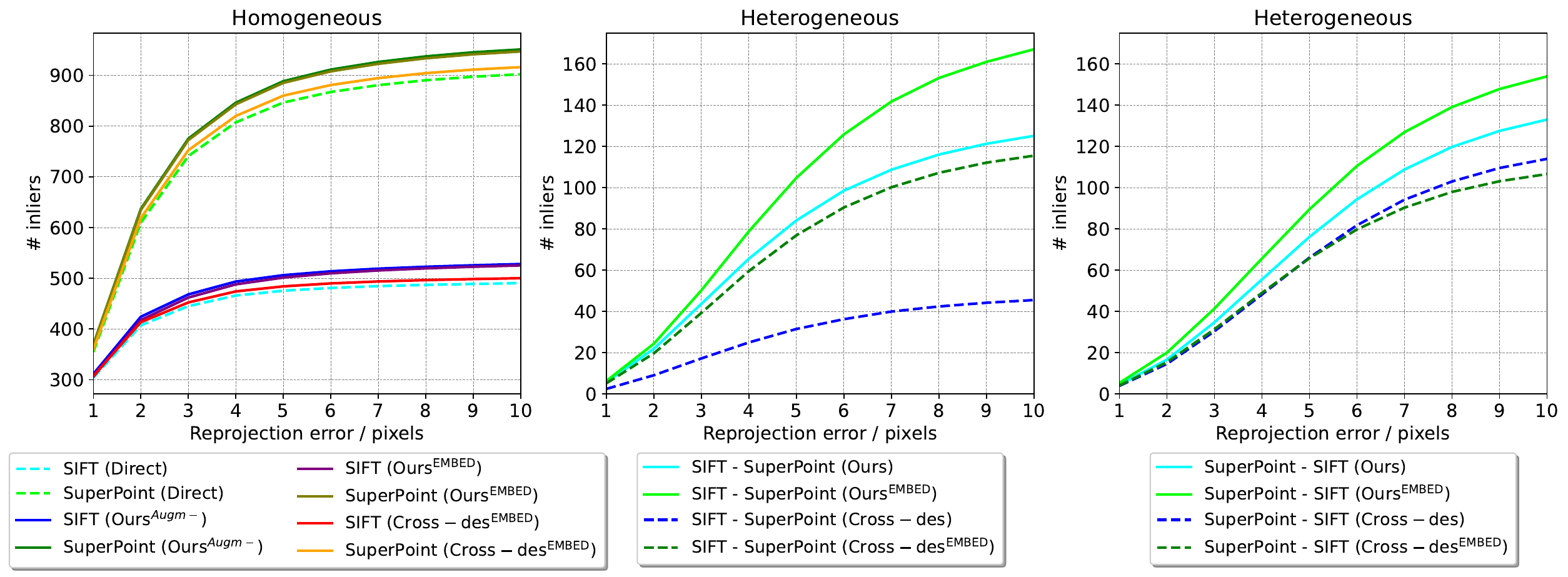}
            \caption[]{\small Results for number of inliers in the HPatches sequences dataset}    
            \label{fig:inliers}
        \end{subfigure}
        \caption[]{\small Performance on the HPatches sequences dataset. We average results on MMA and number of inliers for all the sequences, and compare homogeneous (top and bottom left plots) and heterogeneous cases (middle and right plots).} 
        \label{fig:hpatcheseval}
    \end{figure*}

MMA (\cref{fig:mma}) and number of inliers (\cref{fig:inliers}) are evaluated in the homogeneous and heterogeneous cases. First, there is a clear difference in performance between the homogeneous and the heterogeneous cases. For homogeneous cases, for accurate estimations of up to $3$ pixels of reprojection error, more than $50\%$ of the points can be correctly matched (MMA $> 0.5$), as can be seen in \cref{fig:mma} (left). However, in the heterogeneous cases in \cref{fig:mma} (middle, right), this number drops to less than $10\%$ for the same reprojection error threshold. Our method improves over the baseline, and although the difference is greater at higher error thresholds, our method can increase percent of correct matches up to $10\%$ of the number of points. The set of inliers is drastically reduced from homogeneous to heterogeneous cases (see \cref{fig:inliers}), which is to be expected, due to poor keypoint repeatability across detectors. Our method, however, consistently places these numbers above \num{100}, significantly improving over the baseline \cite{crossdes}.

\subsection{Visual localization}
\begin{table*}[!htbp]
    %\begin{minipage}[t]{0.55\textwidth}
    \captionsetup{width=.75\textwidth}
    \caption{Evaluation on Aachen Day and Night v1.1 benchmark for methods trained on SIFT and SuperPoint. Best results in bold, second best underlined.}
    \begin{center}
    \resizebox{0.7\textwidth}{!}{
    %\resizebox{\columnwidth}{!}{
    \begin{tabular}{ccclcccccccc}
    \toprule
    & \multicolumn{2}{c}{\multirow{2}{*}{Algorithm}} & \multicolumn{1}{c}{\multirow{4}{*}{Method}} && \multicolumn{7}{c}{\% localized queries} \\
    && & && \multicolumn{7}{c}{(\SI{0.25}{\meter}, \ang{2}) \ (\SI{0.5}{\meter}, \ang{5}) \ (\SI{5}{\meter}, \ang{10})} \\
    \cmidrule{2-3} \cmidrule{5-12}
    & Map & Query & && \multicolumn{3}{c}{Day} && \multicolumn{3}{c}{Night} \\
    %& & & &&  & & && (0.25m, 2\circ) & ($\SI{0.5}{\meter}$, 5) & (5m, 10 \circ) \\
    \midrule
    \parbox[t]{5mm}{\multirow{10}{*}{\rotatebox[origin=c]{90}{Homogeneous}}} & \multirow{5}{*}{SIFT} & \multirow{5}{*}{SIFT} & Direct 
                                                            && \textbf{82.2} & \textbf{89.1} & \textbf{93.8} && \textbf{41.4} & \textbf{51.3} & \textbf{66.5} \\
    & & & FB                                    && 77.9 & 86.0 & 90.7 && 30.4 & 40.8 & 52.9 \\
    & & & $\text{Ours}^{A-}$                                              && \underline{79.5} & \underline{87.1} & \underline{91.9} && \underline{35.1} & \underline{45.5} & 56.0 \\
    & & & C-D$^{\text{EMB}}$                 && 77.3 & 85.0 & 90.2 && 26.7 & 38.7 & 49.2 \\
    %& & & Ours$^{\text{EMB}}$                             && 79.2 & 87.1 & 91.7 && 35.1 & 44.0 & 58.6 \\
    & & & Ours$^{\text{EMB}}$                             && 78.5 & 86.7 & 91.1 && 33.0 & 42.4 & \underline{58.1} \\
    \cmidrule{2-12}
    & \multirow{5}{*}{SuperPoint} & \multirow{5}{*}{SuperPoint} & Direct && 85.9 & 92.0 & 96.6 && 64.9 & 80.6 & 93.7 \\
    & & & FB                                                && 86.9 & 92.8 & 96.8 && 67.0 & 82.2 & \textbf{96.3} \\
    & & & Ours$^{A-}$                                                          && \underline{87.4} & \underline{93.2} & \underline{97.0} && \textbf{71.2} & \textbf{85.3} & \textbf{96.3} \\
    & & & C-D$^{\text{EMB}}$                             && 86.5 & \underline{93.2} & 96.5 && \underline{68.1} & \underline{83.8} & 93.7 \\
    & & & Ours$^{\text{EMB}}$                                         && \textbf{88.5} & \textbf{94.1 }& \textbf{97.3} && \textbf{71.2} & 83.2 & \underline{94.8} \\
    %& & & Ours$^{\text{EMB}}$                                         && 87.1 & 93.0 & 97.6 && 67.5 & 82.2 & 94.2 \\
    %& & & & & & & & & & \\
    \midrule
    %& & & & & & & & & & \\
    \parbox[t]{5mm}{\multirow{10}{*}{\rotatebox[origin=c]{90}{Heterogeneous}}} & \multirow{4}{*}{SIFT} & \multirow{4}{*}{SuperPoint} & C-D && 43.4 & 56.3 & 73.9 && 8.9 & 15.2 & 26.7 \\
    & & & Ours                                              && \underline{55.7} & 65.0 & 80.6 &&  16.8 & \underline{29.3} & \underline{58.1} \\
    & & & C-D$^{\text{EMB}}$                 && 54.5 & \underline{67.6} & \underline{81.4} && \underline{17.3} & 23.6 & 41.9 \\
    & & & Ours$^{\text{EMB}}$                             && \textbf{57.6} & \textbf{69.5} & \textbf{84.3} && \textbf{23.0} & \textbf{34.0} & \textbf{60.7} \\
    %& sift-sift & spt-sift &                                                     && 60.2 & 71.5 & 82.0 && 16.8 & 25.1 & 36.1 \\
    %& \textcolor{red}{sift-sift} & \textcolor{red}{spt-sift} & \textcolor{red}{Only boosting} && \textcolor{red}{63.5} & \textcolor{red}{73.1} & \textcolor{red}{84.2} &&  \textcolor{red}{24.6} & \textcolor{red}{33.5} & \textcolor{red}{55.5} \\
    \cmidrule{2-12}
    & \multirow{6}{*}{SuperPoint} & \multirow{6}{*}{SIFT} & C-D && 42.8 & 54.6 & 67.8 && 6.3 & 8.4 & 18.3 \\
    & & & C-D$^{\text{SG}}$                           && \underline{48.3} & \underline{60.9} & \underline{70.6} && 6.8 & 9.9 & 11.5 \\
    & & & C-D$^{\text{LG}}$                           && \textbf{66.4} & \textbf{81.7} & \textbf{89.8} && \textbf{18.8} & \textbf{28.8} & \textbf{47.1} \\
    & & & Ours                                              && 47.0 & 56.7 & 70.0 && 6.3 & 10.5 & 25.1 \\
    & & & C-D$^{\text{EMB}}$                        && 40.3 & 52.5 & 64.3 && 5.8 & 9.4 & 16.2 \\
    & & & Ours$^{\text{EMB}}$                             && 48.1 & 56.2 & 69.3 && \underline{7.9} & \underline{14.1} & \underline{29.3} \\
    %& spt-spt & sift-spt &  && 73.1 & 84.2 & 92.6 && 33.0 & 51.3 & 67.5 \\
    %& \textcolor{red}{spt-spt} & \textcolor{red}{sift-spt} & \textcolor{red}{Only boosting (25)} && \textcolor{red}{73.9} & \textcolor{red}{84.0} & \textcolor{red}{91.0} &&  \textcolor{red}{34.0} & \textcolor{red}{52.9} & \textcolor{red}{74.3} \\
    \bottomrule
    \end{tabular}
    }
    \label{tab:aachen}
    \end{center}
    %\end{minipage}%
    \hfill
%\end{minipage}
\end{table*}

\begin{table*}[!htbp]
    \caption{Evaluation on 7Scenes dataset. Best results in bold.}
    \begin{center}
    \resizebox{0.8\textwidth}{!}{
    \begin{tabular}{ccclcccccccc}
    \toprule
    & \multicolumn{2}{c}{\multirow{2}{*}{Algorithm}} & \multicolumn{1}{c}{\multirow{4}{*}{Method}} && \multicolumn{7}{c}{Median Pose Error} \\
    && & && \multicolumn{7}{c}{(cm, $^\circ$)} \\
    \cmidrule{2-3} \cmidrule{5-12}
    & Map & Query & && Chess & Fire & Heads & Office & Pumpkin & Kitchen & Stairs \\
    %& & & &&  & & && (0.25m, 2\circ) & ($\SI{0.5}{\meter}$, 5) & (5m, 10 \circ) \\
    \midrule
    \parbox[t]{5mm}{\multirow{10}{*}{\rotatebox[origin=c]{90}{Homogeneous}}} & \multirow{4}{*}{SIFT} & \multirow{4}{*}{SIFT} & Direct && 2.50,\,0.88 & 2.02,\,\textbf{0.85} & 1.12,\,0.83 & \textbf{3.47},\,1.04 & 5.57,\,1.54 & \textbf{4.48},\,\textbf{1.49} & \textbf{4.79},\,\textbf{1.26} \\
    & & & FB  &&  2.51,\,\textbf{0.87} & \textbf{2.01},\,\textbf{0.85} & 1.09,\,0.82 & 3.54,\,1.03 & 5.51,\,1.53 & 4.51,\,1.50 & 5.02,\,\textbf{1.26} \\
    & & & \text{Ours}$^{A-}$ && 2.52,\,\textbf{0.87} & \textbf{2.01},\,0.86 & 1.09,\,0.83 & 3.51,\,\textbf{1.02} & 5.55,\,\textbf{1.52} & 4.53,\,1.51 & 4.89,\,\textbf{1.26} \\
    & & & C-D$^{\text{EMB}}$ && \textbf{2.49},\,\textbf{0.87} & 2.04,\,0.87 & 1.09,\,\textbf{0.81} & 3.48,\,1.04 & 5.59,\,1.53 & 4.49,\,1.50 & 4.83,\,1.28 \\
    & & & Ours$^{\text{EMB}}$ &&  2.53,\,0.88 & 2.05,\,0.86 & \textbf{1.07},\,0.83 & 3.55,\,1.04 & \textbf{5.43},\,\textbf{1.52} & 4.52,\,1.51 & 4.86,\,1.27 \\
    \cmidrule{2-12}
    & \multirow{4}{*}{SuperPoint} & \multirow{4}{*}{SuperPoint} & Direct  && \textbf{2.49},\,0.86 & 2.11,\,\textbf{0.86} & 1.14,\,0.81 & 3.42,\,\textbf{1.02} & \textbf{5.60},\,1.48 & \textbf{4.70},\,\textbf{1.53} & 5.71,\,1.54 \\
    & & & FB  && \textbf{2.49},\,0.86 & 2.11,\,\textbf{0.86} & \textbf{1.12},\,0.81 & 3.41,\,\textbf{1.02} & 5.75,\,\textbf{1.46} & 4.71,\,1.54 & 5.56,\,\textbf{1.50} \\
    & & & Ours$^{A-}$ && 2.50,\,\textbf{0.85} & \textbf{2.10},\,\textbf{0.86} & 1.16,\,0.83 & 3.42,\,1.03 & 5.71,\,1.48 & 4.73,\,1.54 & \textbf{5.41},\,1.51 \\
    & & & C-D$^{\text{EMB}}$ && 2.50,\,\textbf{0.85} & 2.13,\,0.87 & 1.14,\,\textbf{0.80} & \textbf{3.40},\,\textbf{1.02} & 5.80,\,1.48 & \textbf{4.70},\,\textbf{1.53} & 5.46,\,1.51 \\
    & & & Ours$^{\text{EMB}}$ && \textbf{2.49},\,\textbf{0.85} & \textbf{2.10},\,\textbf{0.86} & 1.14,\,0.82 & 3.42,\,1.03 & 5.84,\,1.50 & 4.74,\,1.54 & \textbf{5.41},\,\textbf{1.50} \\
    \midrule
    \parbox[t]{5mm}{\multirow{8}{*}{\rotatebox[origin=c]{90}{Heterogeneous}}} & \multirow{4}{*}{SIFT} & \multirow{4}{*}{SuperPoint} & C-D  && 3.45,\,1.04 & 3.11,\,1.23 & 2.08,\,1.42 & 5.07,\,1.43 & 6.96,\,1.95 & 5.67,\,1.77 & 16.28,\,4.25 \\
    & & & Ours && \textbf{2.82},\,\textbf{0.93} & 2.82,\,1.08 & 1.64,\,1.16 & 4.45,\,1.32 & \textbf{6.47},\,\textbf{1.85} & 5.22,\,1.68 & 11.47,\,2.93 \\
    & & & C-D$^{\text{EMB}}$ && 3.18,\,0.95 & \textbf{2.78},\,\textbf{1.03} & \textbf{1.60},\,\textbf{1.14} & 4.47,\,\textbf{1.29} & 6.58,\,1.86 & \textbf{5.20},\,\textbf{1.64} & 9.95,\,2.67 \\
    & & & Ours$^{\text{EMB}}$ && 2.85,\,0.94 & 2.81,\,1.09 & 1.69,\,1.20 & \textbf{4.42},\,1.31 & 6.71,\,1.86 & 5.22,\,1.66 & \textbf{9.10},\,\textbf{2.43} \\
    \cmidrule{2-12}
    & \multirow{4}{*}{SuperPoint} & \multirow{4}{*}{SIFT} & C-D  && 3.08,\,1.00 & 3.04,\,1.11 & 2.29,\,1.46 & 5.00,\,1.34 & 7.68,\,1.90 & 5.80,\,1.65 & 10.86,\,2.87 \\
    & & & Ours && 2.90,\,0.96 & 3.07,\,1.18 & 2.02,\,1.46 & 4.76,\,1.37 & \textbf{7.30},\,\textbf{1.81} & 5.52,\,1.67 & 11.78,\,2.96 \\
    & & & C-D$^{\text{EMB}}$ &&  3.01,\,0.96 & \textbf{3.03},\,\textbf{1.10} & 2.15,\,1.42 & 4.94,\,\textbf{1.32} & 7.49,\,1.85 & 5.71,\,\textbf{1.64} & \textbf{9.89},\,\textbf{2.71} \\
    & & & Ours$^{\text{EMB}}$ &&  \textbf{2.83},\,\textbf{0.94} & 3.18,\,1.22 & \textbf{1.96},\,\textbf{1.39} & \textbf{4.67},\,1.35 & 7.41,\,1.84 & \textbf{5.46},\,1.65 & 10.68,\,\textbf{2.71} \\
    \bottomrule
    \end{tabular}
    }
    \label{tab:7svl}
    \end{center}
\end{table*}

Results of visual localization on the Aachen benchmark are shown in \cref{tab:aachen}.    
As in the previous task, the performance gap between homogeneous to heterogeneous cases is very significant. Performance is in general at least halved when moving to cross-algorithm cases, with the results for night queries being in particular sensitive to using different detectors and descriptors. Our method, however, shows consistently better performance than the baseline.
For different map-query algorithm combinations, localization error thresholds and day/night splits our method increases the cross-algorithm performance significantly when using traditional matchers. The baseline \textit{C-D} with deep matchers, $^\text{LG}$ and $^\text{SG}$ shows better performance than our method, suggesting the need to continue research for cross-algorithm scenarios towards deep matchers.
\cref{tab:7svl} shows the median errors in 7-Scenes, where the gap between homogeneous and heterogeneous algorithms is not as big, as is indoors, and scenarios are simple and small (i.e., non-planar Stairs scene worsens notably). Performance in the homogeneous case is paired for all methods, but our method equals or outperforms the baseline mostly in the heterogeneous scenarios. Note that our method, seeking matching improvement in cross-algorithm scenarios, does not compromise the performance for homogeneous cases, but it is expected to also not improve significantly over direct methods, since that's not the main objective of the augmentation networks. %Expanding our method to include one more feature in \cref{tab:aachendisk}, our method shows minimal decrease in performance for previously tested configurations. In this case, however, our performance is tied to the baseline, which might indicate the need to use SuperPoint as a detector as well when augmenting DISK to increase performance beyond these values.

\section{Discussion}
%Repeatability
Our work results in improved performance in cross-detector scenarios over the baseline work. The main limitation comes from relying on the repeatability of keypoints detected cross-algorithm. To obtain an estimate of a relative pose, a set of correspondences of size $\geqslant 5$ is in general needed. Given noise and outliers, this set should be considerably larger, which can not be guaranteed if two different detectors are used. Future work includes refining geometric characteristics of keypoints, including their $x$ and $y$ coordinates, for the cross-algorithm matching problem. Furthermore, given the improvement that LightGlue \cite{lightglue} presents over NN for matching, future work will consider integration of our embedded space descriptors with deep matchers.
% Scalability
%As \cite{crossdes} showed, performance is not decreased significantly from having one encoder and decoder to translate in between pairs of feature descriptors compared to several encoders-decoders trained together for one same embedded space. However, it is still not clear how scalable this approach is, as in practice a variety of feature detection algorithms are used in practice. 

\section{Conclusions}
Our paper presents a first approach to decreasing the cross-detector performance gap in the state-of-the-art, with our method presenting a solution that is specific to cross-algorithm scenarios. We highlight the difficulty of the heterogeneous algorithm problem by exploring the drastic performance drop when moving from homogeneous to heterogeneous cases. We show how our method, introducing a detector-aware descriptor augmentation, significantly improves results when compared to SotA for image matching and visual localization tasks.

%\section*{Acknowledgements}
% WASP blabla. Need to leave it empty for the blind submission?

\section*{Acknowledgements}
This work was partially supported by the Wallenberg AI, Autonomous Systems and Software Program (WASP) funded by the Knut and Alice Wallenberg Foundation.

\newpage\phantom{blabla}
\newpage\phantom{blabla}

{\small
\bibliographystyle{ieee_fullname}
\bibliography{main}
}

\newpage
\clearpage
\setcounter{page}{1}
\maketitlesupplementary
\makeatletter
\renewcommand \thesection{\Alph{section}}
\setcounter{section}{0}
\renewcommand*{\theHsection}{chX.\the\value{section}}

\renewcommand\thetable{\Alph{table}}
\setcounter{table}{0}
\renewcommand*{\theHtable}{chX.\the\value{table}}

\renewcommand \thefigure{\Alph{figure}}
\setcounter{figure}{0}
\renewcommand*{\theHfigure}{chX.\the\value{figure}}
\makeatother

%\section{Additional architecture details}

%\begin{figure*}[!htbp]
%    \centering
%    \includestandalone[width=\textwidth]{diagrams/extended_architecture}
%    \caption{Architecture.}
%    \label{fig:additionalarchitecture}
%\end{figure*}

\section{Inference time statistics}
We separately measure both the time to augment and to translate to embedded space, for features SIFT and SuperPoint. We use a desktop computer with an Intel Core i9-13900KF CPU and an NVIDIA GeForce RTX 4090 GPU. We perform each measurement for 1000 different images and record statistics on average number of features extracted on each image and inference time per image. Results can be seen in table \cref{tab:time} and show feasibility of executing our method in real-time systems.

    \begin{table}[h]
        \caption{Mean number of features extracted and inference time statistics (mean $\pm$ std. dev.) per image.}
        \resizebox{0.95\columnwidth}{!}{
        \begin{tabular}{r|cc}
        \toprule
        Feature & SIFT & SuperPoint \\%& DISK \\
        \hline
        Mean features & $2046$ & $1996$ \\%& $2048$ \\ 
        Augmentation (\si{\milli\second}) & $9.8 \pm 4$ & $11.4 \pm 4$ \\%& $9.8 \pm 4$ \\
        Translation (\si{\milli\second}) & $ 0.9 \pm 1.4 $ & $1.1 \pm 1.4 $ \\%& $ 0.9 \pm 1.4 $ \\
        \bottomrule
        \end{tabular}
        }
        \label{tab:time}
    \end{table}

%We repeat each measurement 100 times and we find that a forward pass on average takes $14.88 \pm 0.75\,$ms on CPU and $2.26 \pm 0.10\,$ms on GPU, that is, our pipeline can run in real time.

\section{Additional visual localization benchmarking results}
For this experiment, we use the same setup as described in \cref{parag:expsetup_vl}, however, we now arrange a pure cross-detector setup. 
%We detach DoG and SuperPoint detection, and interchangeably describe such keypoints with SIFT and SuperPoint, to obtain features that are always described with the same descriptor algorithm, but have been detected heterogeneously for map and query. 
To obtain features that are always described with the same descriptor algorithm but have been detected heterogeneously for map and query, we detach DoG and SuperPoint detection, and interchangeably describe such keypoints with SIFT and SuperPoint.
The purpose of this experiment is twofold: first, it serves as an ablation study for the relevance of the embedded space when matching across detectors. Second, it provides an upper bound of performance for cross-algorithm scenarios solely due to the impact of having keypoints extracted from different feature detectors.

\begin{table*}[!htbp]
    \caption{Extended evaluation on Aachen Day and Night v1.1 benchmark. The detector and descriptor algorithms are combined to obtain pure cross-detector scenarios, and the specific configuration for map and query feature extraction is explicitly mentioned.}
    \begin{center}
    \resizebox{0.95\textwidth}{!}{
    \begin{tabular}{ccccclcccccccc}
    \toprule
    & \multicolumn{4}{c}{\multirow{2}{*}{Algorithm}} & \multicolumn{1}{c}{\multirow{4}{*}{Method}} && \multicolumn{7}{c}{\% localized queries} \\
    &&&& & && \multicolumn{7}{c}{(\SI{0.25}{\meter}, \ang{2}) \ (\SI{0.5}{\meter}, \ang{5}) \ (\SI{5}{\meter}, \ang{10})} \\
    \cmidrule{2-5} \cmidrule{7-14}
    & \multicolumn{2}{c}{Map} & \multicolumn{2}{c}{Query} & && \multicolumn{3}{c}{Day} && \multicolumn{3}{c}{Night} \\
    %& & & &&  & & && (0.25m, 2\circ) & ($\SI{0.5}{\meter}$, 5) & (5m, 10 \circ) \\
    \midrule
    %\parbox[t]{5mm}{\multirow{1}{*}{\rotatebox[origin=c]{90}{Het. detection}}}
    & DoG  & SIFT & SuperPoint & SIFT & Direct  && 60.2 & 71.5 & 82.0 && 16.8 & 25.1 & 36.1 \\
    & DoG  & SIFT & SuperPoint & SIFT & Ours$^{A-}$ && 63.5 & \textbf{73.1} & 84.2 && \textbf{24.6} & 33.5 & 55.5 \\
    & DoG  & SIFT & SuperPoint & SIFT & Ours$^{\text{EMB}}$ && \textbf{63.7} & \textbf{73.1} & \textbf{86.0} && 24.1 & \textbf{36.6} & \textbf{63.4} \\
    \cmidrule{2-14}
    & DoG  & SuperPoint & SuperPoint & SuperPoint & Direct  && 65.5 & 78.6 & 88.6 && 32.5 & 51.3 & 73.8 \\
    & DoG  & SuperPoint & SuperPoint & SuperPoint & Ours$^{A-}$ && \textbf{74.8} & \textbf{86.0} & \textbf{94.1} && 53.9 & 72.8 & 85.3 \\
    & DoG  & SuperPoint & SuperPoint & SuperPoint & Ours$^{\text{EMB}}$ && 74.6 & 85.1 & 93.0 && \textbf{55.5} & \textbf{73.3} & \textbf{86.4} \\
    \midrule
    & SuperPoint & SIFT & DoG & SIFT & Direct  && 15.2 & 20.5 & 26.9 && 2.6 & 5.8 & 8.9 \\
    & SuperPoint & SIFT & DoG & SIFT & Ours$^{A-}$  && 29.4 & 36.8 & 46.6 && \textbf{5.2} & 6.8 & 12.6 \\
    & SuperPoint & SIFT & DoG & SIFT & Ours$^{\text{EMB}}$  && \textbf{34.6} & \textbf{41.4} & \textbf{51.2} && \textbf{5.2} & \textbf{10.5} & \textbf{18.3} \\
    \cmidrule{2-14}
    & SuperPoint & SuperPoint & DoG & SuperPoint & Direct  && 73.1 & \textbf{84.2} & \textbf{92.6} && 33.0 & 51.3 & 67.5 \\
    & SuperPoint & SuperPoint & DoG & SuperPoint & Ours$^{A-}$  && \textbf{73.9} & 84.0 & 91.0 && \textbf{34.0} & \textbf{52.9} & \textbf{74.3} \\
    & SuperPoint & SuperPoint & DoG & SuperPoint & Ours$^{\text{EMB}}$  && 73.3 & 83.4 & 91.6 && 30.9 & 46.1 & 72.8 \\
    
    %& DoG & SIFT & DoG & SIFT & Direct && 82.2 & 89.1 & 93.8 && 41.4 & 51.3 & 66.5 \\
    %& SuperPoint & SIFT & SuperPoint & SIFT & Direct  &&  &  &  &&  &  &  \\
    %& DoG & SuperPoint & DoG & SuperPoint & Direct  && 78.8 & 87.7 & 92.6 && 44.0 & 58.1 & 76.4 \\
    %& SuperPoint & SuperPoint & SuperPoint & SuperPoint & Direct  && 85.9 & 92.0 & 96.6 && 64.9 & 80.6 & 93.7 \\
    %& & & & & & & & & & \\
    %& \textcolor{red}{sift}& sift & \textcolor{red}{spt} & sift & \textcolor{red}{Only boosting} && \textcolor{red}{63.5} & \textcolor{red}{73.1} & \textcolor{red}{84.2} &&  \textcolor{red}{24.6} & \textcolor{red}{33.5} & \textcolor{red}{55.5} \\
    %& \textcolor{red}{spt} & spt & \textcolor{red}{sift} & spt & \textcolor{red}{Only boosting (25)} && \textcolor{red}{73.9} & \textcolor{red}{84.0} & \textcolor{red}{91.0} &&  \textcolor{red}{34.0} & \textcolor{red}{52.9} & \textcolor{red}{74.3} \\
    \bottomrule
    \end{tabular}
    }
    \label{tab:aachen2}
    \end{center}
\end{table*}

Results are shown on \cref{tab:aachen2}. It can be seen how, for most configurations, our method performs best when adding the cross-algorithm feature translation contribution (detailed in \cref{ssec:translation}), which encodes augmented descriptors to an embedded space, and using the embedded space as the matching space for features. This can be explained by the higher dimensionality of the space, which could suit better the detector-aware augmented descriptors (detailed in \cref{ssec:dada}), accompanied by the fact that the cross-algorithm feature translation further encodes features to guarantee more discriminative descriptors.

Results in \cref{tab:aachen2} further show the impact of heterogeneous detection algorithms for visual localization. For example, visual localization is sensitive to a specific feature configuration: map detected and described with DoG and SuperPoint, and queries detected and described with DoG and SIFT. This can stem from the fact that SIFT is not a suitable descriptor for keypoints extracted with other algorithms than DoG and similar. Nonetheless, our method greatly improves results for the day queries with respect to the original results with the direct method by simply performing Nearest Neighbors (NN) and a mutual check.

\section{Inlier visualization for the visual localization task}
\begin{figure*}[!htbp]
        \centering
        \begin{subfigure}[b]{\textwidth}
            \centering           
            \includegraphics[width=0.92\textwidth]{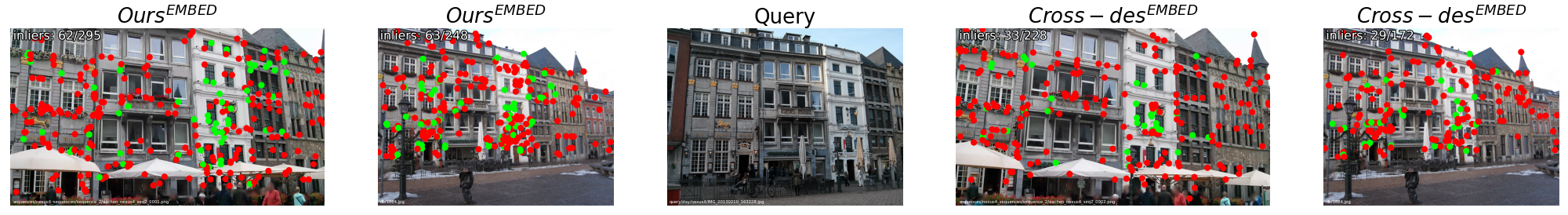}
            \caption[]{\footnotesize Localization inliers for \texttt{query/day\_nexus4/IMG\_20130210\_163228.jpg}}
            \label{fig:sift-spt-2}
        \end{subfigure}
        \begin{subfigure}[b]{\textwidth}
            \centering           
            \includegraphics[width=0.85\textwidth]{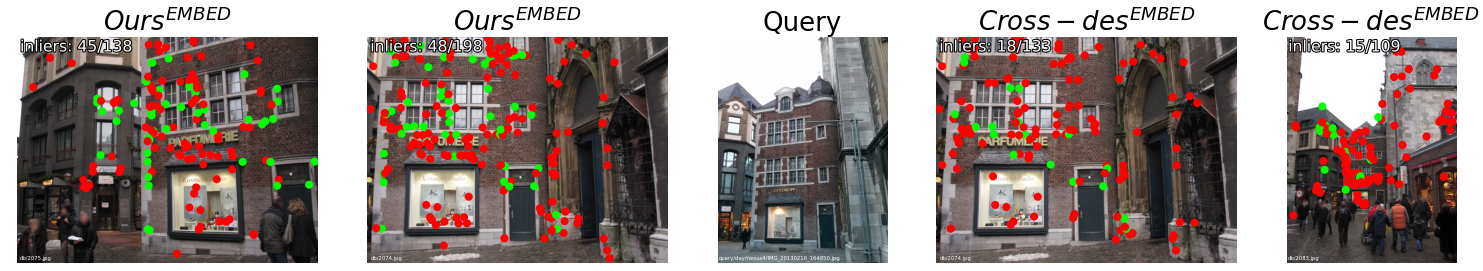}
            %\caption[]{{\small }}    
            \caption[]{\footnotesize Localization inliers for \texttt{query/day\_nexus4/IMG\_20130210\_164850.jpg}}
            \label{fig:sift-spt-3}
        \end{subfigure}
        \begin{subfigure}[b]{\textwidth}
            \centering           
            \includegraphics[width=0.90\textwidth]{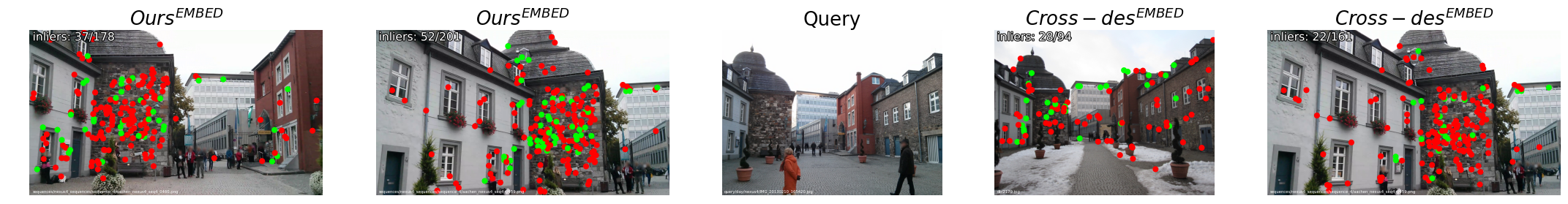}
            \caption[]{\footnotesize Localization inliers for \texttt{query/day\_nexus4/IMG\_20130210\_165420.jpg}}
            \label{fig:sift-spt-4}
        \end{subfigure}
        \begin{subfigure}[b]{\textwidth}
            \centering           
            \includegraphics[width=0.90\textwidth]{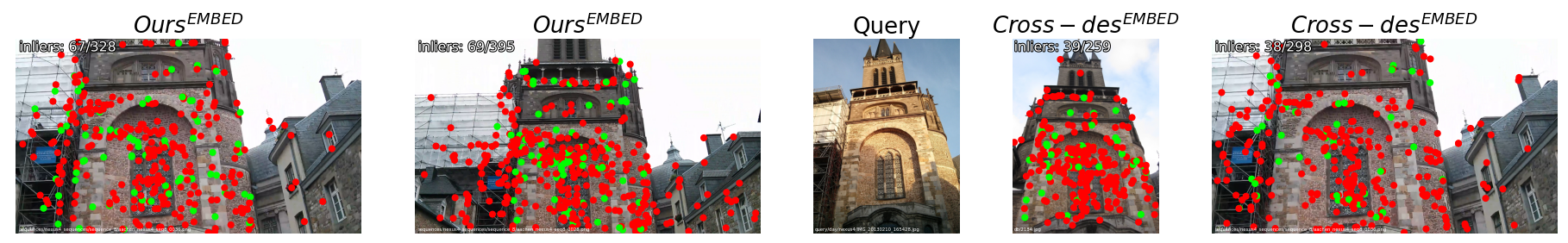}
            \caption[]{\footnotesize Localization inliers for \texttt{query/day\_nexus4/IMG\_20130210\_165428.jpg}}
            \label{fig:sift-spt-5}
        \end{subfigure}
        \begin{subfigure}[b]{\textwidth}
            \centering           
            \includegraphics[width=0.82\textwidth]{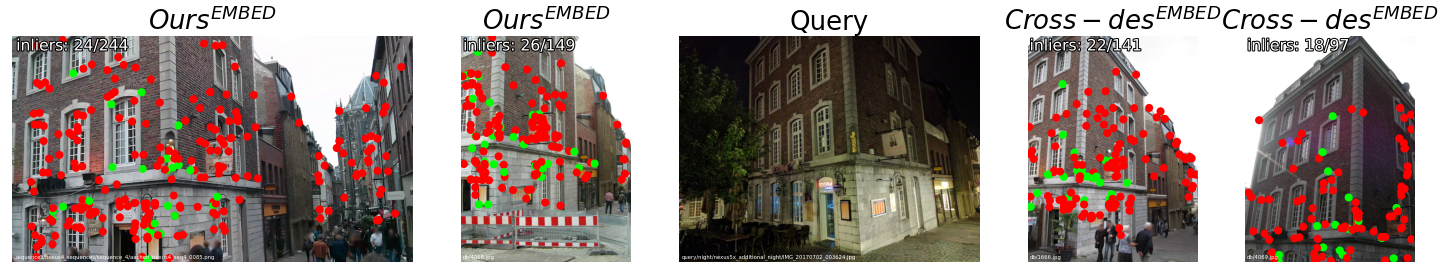}
            \caption[]{\footnotesize Localization inliers for \texttt{query/night\_nexus5x\_additional\_night/IMG\_20170702\_003624.jpg}}
            \label{fig:sift-spt-8}
        \end{subfigure}

        \begin{subfigure}[b]{\textwidth}
            \centering           
            \includegraphics[width=0.87\textwidth]{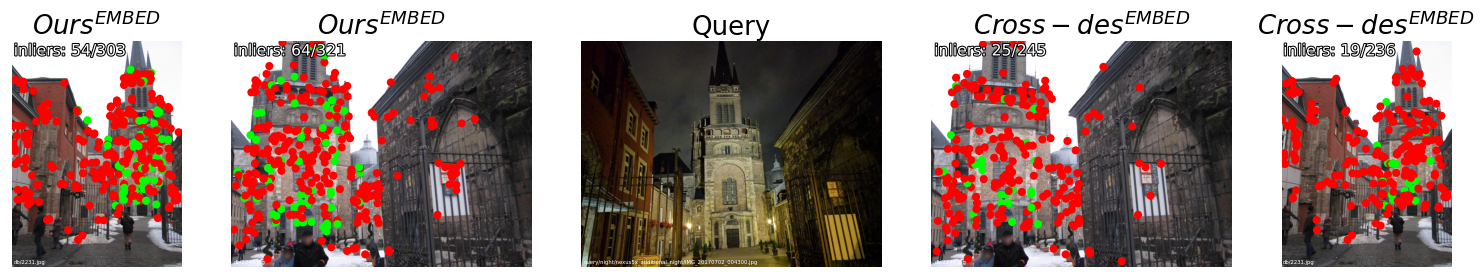}
            \caption[]{\footnotesize Localization inliers for \texttt{query/night\_nexus5x\_additional\_night/IMG\_20170702\_004300.jpg}}
            \label{fig:sift-spt-10}
        \end{subfigure}
        \begin{subfigure}[b]{\textwidth}
            \centering           
            \includegraphics[width=0.90\textwidth]{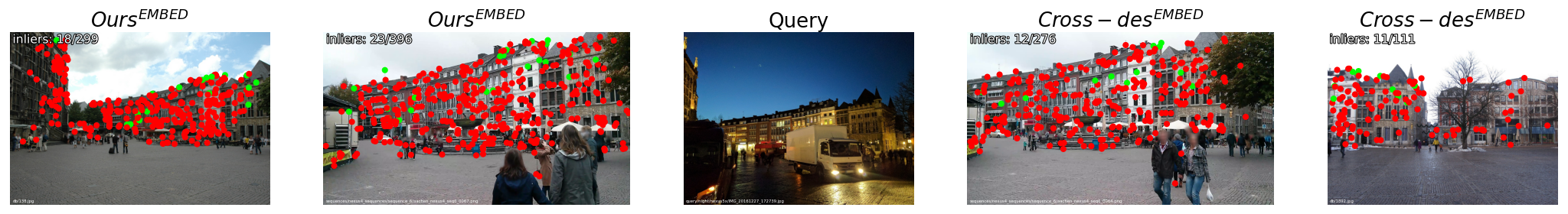}
            \caption[]{\footnotesize Localization inliers for \texttt{query/night\_nexus5x/IMG\_20161227\_172739.jpg}}
            \label{fig:sift-spt-11}
        \end{subfigure}
        \caption[]{\small Qualitative results on the Aachen Day and Night v1.1 benchmark \cite{aachen}. \textbf{Map} feature extraction with \textbf{SIFT}, \textbf{query} feature extraction with \textbf{SuperPoint}. The top-two images with largest inlier set from the reference map are shown for a given query image. Inliers (green) against the total number matches (sum of inliers (green) and outliers (red) are displayed in the top left corner of the retrieved map images.} 
        \label{fig:sift-spt-viz2}
\end{figure*}

\begin{figure*}[!htbp]
        \centering
        %\begin{subfigure}[b]{0.95\textwidth}
        %    \centering           
        %    \includegraphics[width=\textwidth]{images/superpoint_sift_viz2/query_day_nexus4_IMG_20130210_163213_508_matching.png}
        %    %\caption[]{{\small }}    
        %    \label{fig:spt-sift-1}
        %\end{subfigure}
        \begin{subfigure}[b]{\textwidth}
            \centering           
            \includegraphics[width=\textwidth]{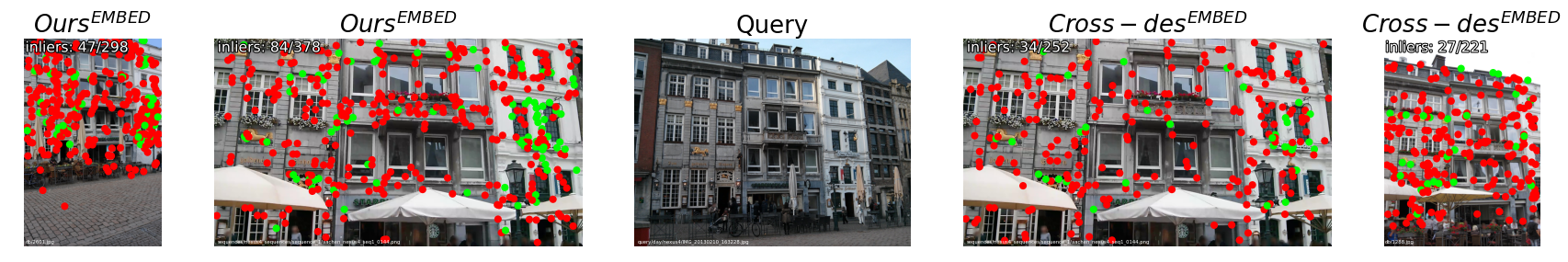}
            \caption[]{\footnotesize Localization inliers for \texttt{query/day\_nexus4/IMG\_20130210\_163228.jpg}}
            \label{fig:spt-sift-2}
        \end{subfigure}
        \begin{subfigure}[b]{\textwidth}
            \centering           
            \includegraphics[width=0.92\textwidth]{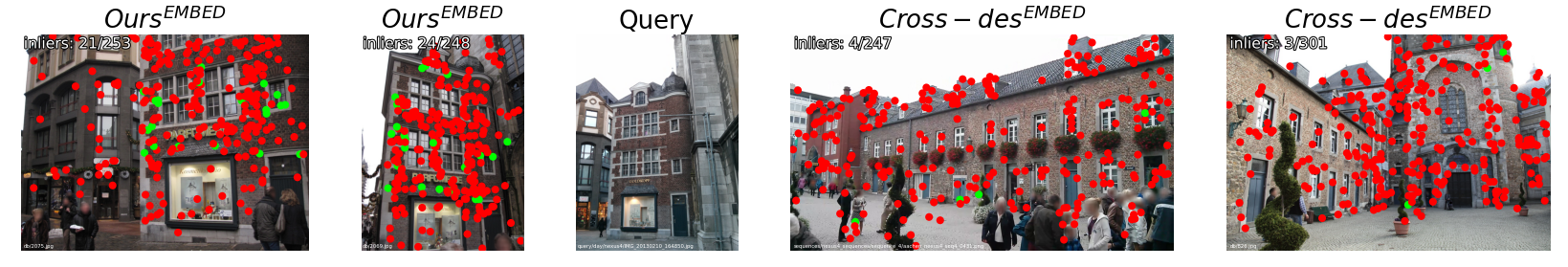}
            %\caption[]{{\small }}    
            \caption[]{\footnotesize Localization inliers for \texttt{query/day\_nexus4/IMG\_20130210\_164850.jpg}}
            \label{fig:spt-sift-3}
        \end{subfigure}
        \begin{subfigure}[b]{\textwidth}
            \centering           
            \includegraphics[width=\textwidth]{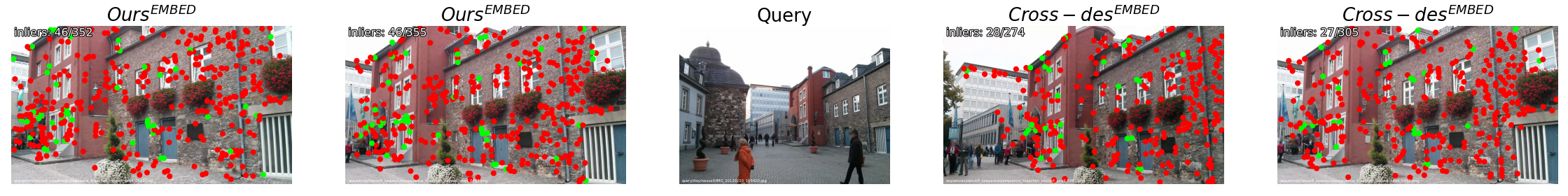}
            \caption[]{\footnotesize Localization inliers for \texttt{query/day\_nexus4/IMG\_20130210\_165420.jpg}}
            \label{fig:spt-sift-4}
        \end{subfigure}
        \begin{subfigure}[b]{\textwidth}
            \centering           
            \includegraphics[width=0.92\textwidth]{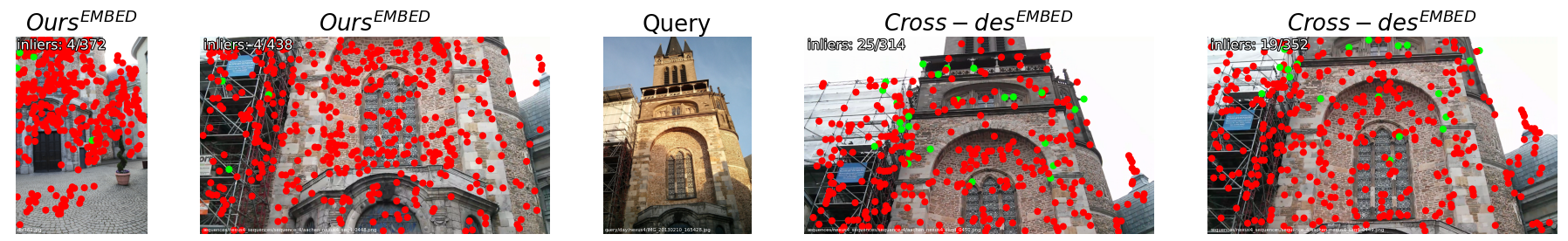}
            \caption[]{\footnotesize Localization inliers for \texttt{query/day\_nexus4/IMG\_20130210\_165428.jpg}}
            \label{fig:spt-sift-5}
        \end{subfigure}
        %\begin{subfigure}[b]{\textwidth}
        %    \centering           
        %    \includegraphics[width=\textwidth]{images/superpoint_sift_viz2/query_day_nexus4_IMG_20130210_165733_367_matching.png}
        %    %\caption[]{{\small }}    
        %    \label{fig:spt-sift-6}
        %\end{subfigure}
        %\begin{subfigure}[b]{\textwidth}
        %    \centering           
        %    \includegraphics[width=\textwidth]{images/superpoint_sift_viz2/query_day_nexus4_IMG_20130210_165818_426_matching.png}
        %    %\caption[]{{\small }}    
        %    \label{fig:spt-sift-7}
        %\end{subfigure}
        \begin{subfigure}[b]{\textwidth}
            \centering           
            \includegraphics[width=\textwidth]{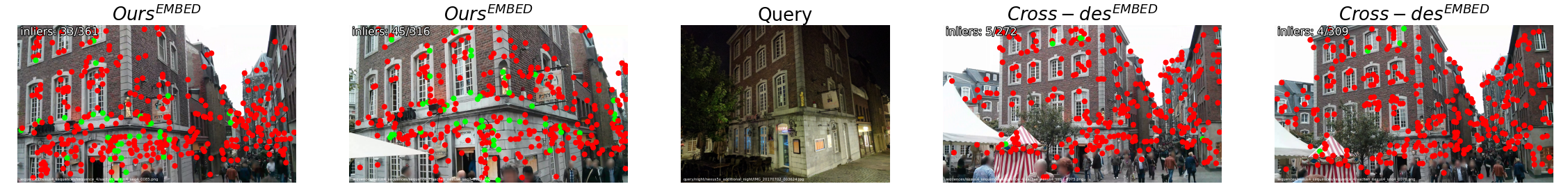}
            \caption[]{\footnotesize Localization inliers for \texttt{query/night\_nexus5x\_additional\_night/IMG\_20170702\_003624.jpg}}
            \label{fig:spt-sift-8}
        \end{subfigure}
        %\begin{subfigure}[b]{\textwidth}
        %    \centering           
        %    \includegraphics[width=\textwidth]{images/superpoint_sift_viz2/query_night_nexus5x_additional_night_IMG_20170702_004219_154_matching.png}
        %    %\caption[]{{\small }}    
        %    \label{fig:spt-sift-9}
        %\end{subfigure}
        \begin{subfigure}[b]{\textwidth}
            \centering           
            \includegraphics[width=0.98\textwidth]{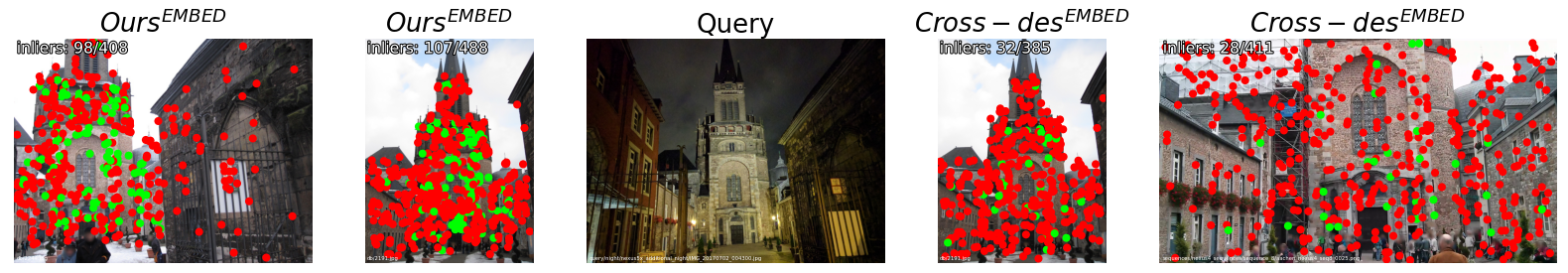}
            \caption[]{\footnotesize Localization inliers for \texttt{query/night\_nexus5x\_additional\_night/IMG\_20170702\_004300.jpg}}
            \label{fig:spt-sift-10}
        \end{subfigure}
        \begin{subfigure}[b]{\textwidth}
            \centering           
            \includegraphics[width=\textwidth]{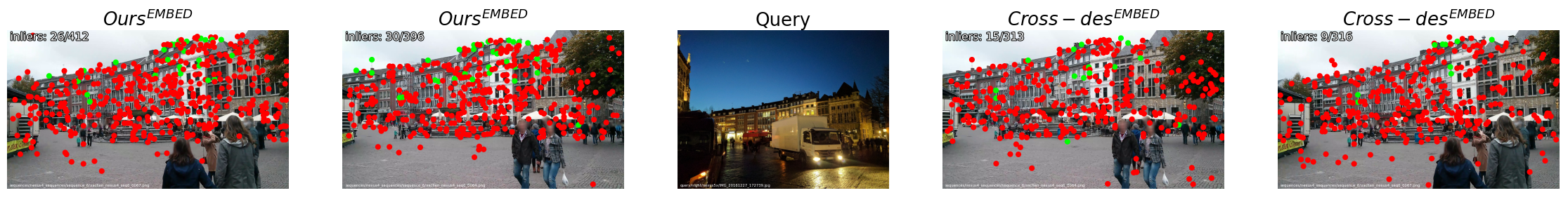}
            \caption[]{\footnotesize Localization inliers for \texttt{query/night\_nexus5x/IMG\_20161227\_172739.jpg}}
            \label{fig:spt-sift-11}
        \end{subfigure}
        %\begin{subfigure}[b]{\textwidth}
        %   \centering           
        %    \includegraphics[width=\textwidth]{images/superpoint_sift_viz2/query_night_nexus5x_IMG_20161227_191808_34_matching.png}
        %    %\caption[]{{\small }}    
        %    \label{fig:spt-sift-12}
        %\end{subfigure}
        %\begin{subfigure}[b]{\textwidth}
        %    \centering           
        %    \includegraphics[width=\textwidth]{images/superpoint_sift_viz2/query_night_nexus5x_IMG_20161227_191819_21_matching.png}
        %    %\caption[]{{\small }}    
        %    \label{fig:spt-sift-13}
        %\end{subfigure}
        %\begin{subfigure}[b]{\textwidth}
        %    \centering           
        %    \includegraphics[width=\textwidth]{images/superpoint_sift_viz2/query_night_nexus5x_IMG_20161227_191835_25_matching.png}
        %    %\caption[]{{\small }}    
        %    \label{fig:spt-sift-14}
        %\end{subfigure}
        %\begin{subfigure}[b]{\textwidth}
        %    \centering           
        %    \includegraphics[width=\textwidth]{images/superpoint_sift_viz2/query_night_nexus5x_IMG_20161227_191920_71_matching.png}
        %    %\caption[]{{\small }}    
        %    \label{fig:spt-sift-15}
        %\end{subfigure}
        \caption[]{\small Qualitative results on the Aachen Day and Night v1.1 benchmark \cite{aachen}. \textbf{Map} feature extraction with \textbf{SuperPoint}, \textbf{query} feature extraction with \textbf{SIFT}. The top-two images with largest inlier set from the reference map are shown for a given query image. Inliers (green) against the total number matches (sum of inliers (green) and outliers (red)) are displayed in the top left corner of the retrieved map images.} 
        \label{fig:spt-sift-viz2}
\end{figure*}

For a given query image in the Aachen Day and Night v1.1 dataset \cite{aachen}, we plot the two top matched images with the biggest number of inliers in the reference map for our method and against the baseline \cite{crossdes}. Same experimental setup applies as in \cref{parag:expsetup_vl}, and we consider SIFT and SuperPoint as our features. We visualize matches for the cross-algorithm cases, where we extracted SIFT features (DoG detection and SIFT description) for the map images, and SuperPoint features (SuperPoint algorithm for detection and description coupled with S3Esti \cite{s3esti}) for the query images in \cref{fig:sift-spt-viz2}; and vice versa in \cref{fig:spt-sift-viz2}.

Our method consistently identifies a larger number of inliers, even for queries taken at night. Given cross-detector scenarios, both for the baseline and our method, the number of inliers is small relative to the number of features matched with NN and mutual check, sometimes even two orders of magnitude smaller. Hence, it is extremely important to expand the inlier set, which our method consistently allows over the baseline. In \cref{fig:spt-sift-11} both methods find the same top-two images in the map for the given query, and results can be directly compared. Both methods identify correctly matches in the top of the white building, however our method can identify more features in the same area and also in more challenging areas. There are other examples, however, where our method is outperformed by the baseline, for example in \cref{fig:spt-sift-5}, where our method can at maximum identify four inliers, whereas the baseline retrieves a bigger set. However, with the opposite feature combination for map and queries, as shown in \cref{fig:sift-spt-5}, our method outperforms again the baseline for the same query image.

\clearpage

\clearpage

\end{document}